\pdfoutput=1

\documentclass[11pt]{article}

\usepackage{EMNLP2023}

\usepackage{times}
\usepackage{latexsym}

\usepackage[T1]{fontenc}

\usepackage[utf8]{inputenc}

\usepackage{microtype}
\usepackage{inconsolata}
\usepackage{caption}
\usepackage{subcaption}
\usepackage{graphicx}
\usepackage{algorithm} 
\usepackage{algpseudocode} 
\usepackage{amsmath}
\usepackage{enumerate}
\usepackage{enumitem}
\usepackage[english]{babel} 
\usepackage{blindtext}
\usepackage{multicol}
\usepackage{multirow}
\usepackage{booktabs}
\usepackage{tikz}
\usepackage{ulem}
\usepackage{arydshln}

\DeclareMathOperator*{\argmax}{arg\,max}

\usepackage[linecolor=orange,size=scriptsize]{todonotes}

\newcommand\Tstrut{\rule{0pt}{2.6ex}}       
\newcommand\Bstrut{\rule[-0.9ex]{0pt}{0pt}} 
\newcommand{\TBstrut}{\Tstrut\Bstrut} 

\newcommand{\Ra}{\texttt{RATION} }

\title{Rationale-based Opinion Summarization}

\author{Haoyuan Li \qquad Snigdha Chaturvedi \\ \texttt{\{haoyuanl, snigdha\}@cs.unc.edu}\\ UNC Chapel Hill}

\begin{document}
\maketitle
\begin{abstract}
Opinion summarization aims to generate concise summaries that present popular opinions of a large group of reviews. However, these 
summaries can be too generic and lack supporting details. To address these issues, we propose a new paradigm for summarizing reviews, rationale-based opinion summarization. Rationale-based opinion summaries output the representative opinions as well as one or more corresponding rationales. To extract good rationales, we define four desirable properties: 
relatedness, specificity, popularity, and diversity and present a Gibbs-sampling-based method to extract rationales. 
Overall, we propose \Ra, an unsupervised extractive system that has two components: an Opinion Extractor (to extract representative opinions) and Rationales Extractor (to extract corresponding rationales). We conduct automatic and human evaluations to show that rationales extracted by \Ra have the proposed properties and its 
summaries are more useful than conventional summaries. The implementation of our work is available at \href{https://github.com/leehaoyuan/RATION}{https://github.com/leehaoyuan/RATION}

\end{abstract}

\section{Introduction}
\label{intro}
 Online reviews are useful for both customers and businesses~\cite{cheung2012review}. However, the large number of reviews on such platforms makes it difficult 
 to manually read all of them. Opinion summarization aims to tackle this problem by generating a concise summary of the reviews. Recently, much progress has been made in opinion summarization, especially unsupervised summarization. These works either extract sentences from reviews as summaries \cite{zhao2020weakly, angelidis2021extractive,basu-roy-chowdhury-etal-2022-unsupervised,basu-roy-chowdhury-etal-2023-unsupervised-opinion, li2023aspect} or generate summaries conditioned on reviews \cite{chu2019meansum, amplayo-lapata-2020-unsupervised}. 
 However, such summaries are usually very generic and lack supporting evidence.
To address this issue, \citet{suhara2020opiniondigest, bar-haim-etal-2021-every, hosking-etal-2023-attributable} produce summaries in which the summarizing content is attributed to a group of supporting 
review sentences. However, since their goal is to explain the choice of the summary content, the sizes of these groups of supporting sentences are too large to be useful for user consumption.

\begin{figure}[t]
\centering
\includegraphics[width=0.48\textwidth, keepaspectratio]{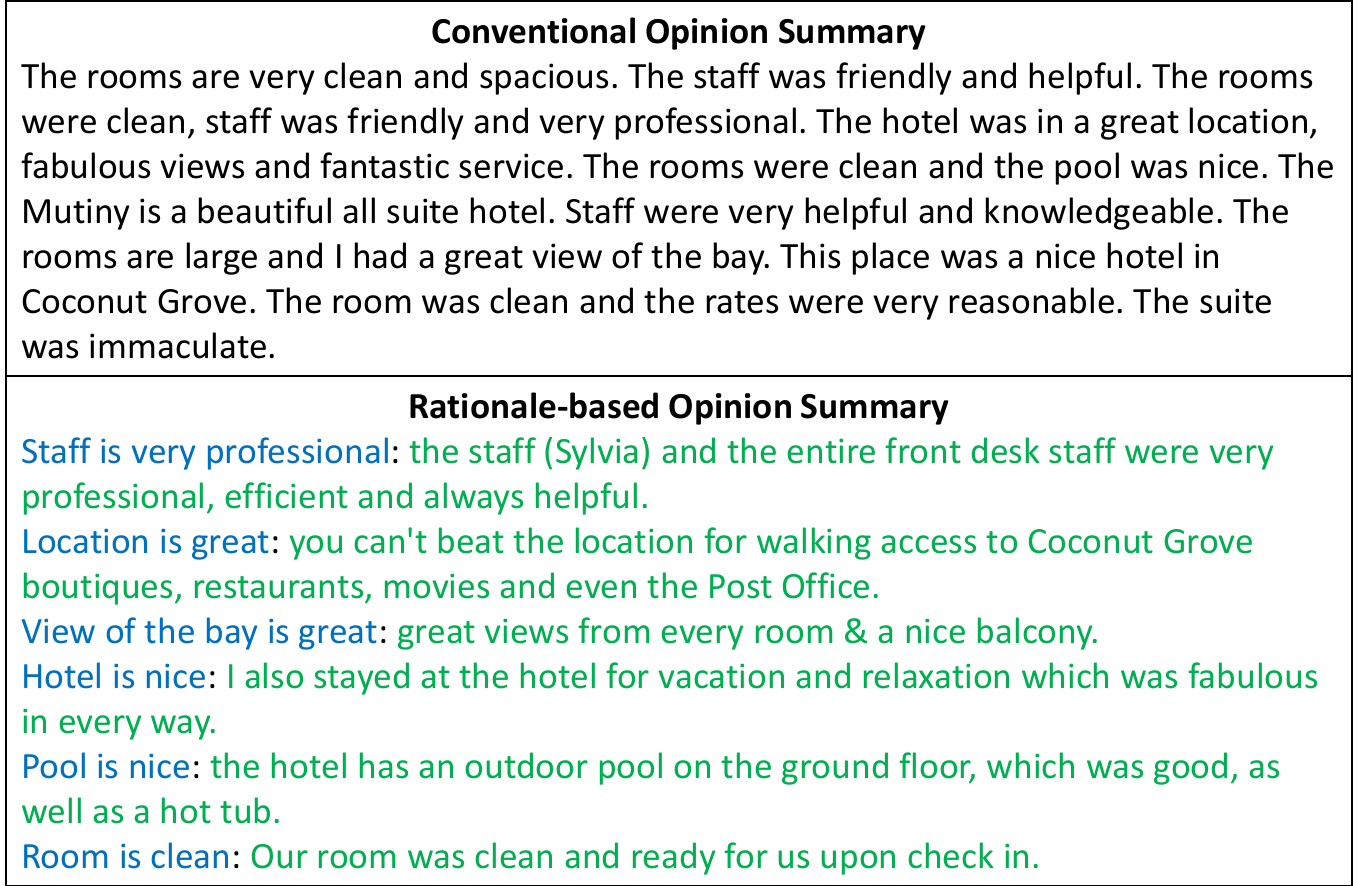}
\caption{Examples of a conventional 
and a rationale-based opinion summary (generated by \Ra) for the same entity. In rationale-based summary, each line presents a \textcolor[RGB]{0,112,192}{representative opinion} and its \textcolor[RGB]{0,176,80}{rationale}. }
\label{fig:compare_sum}
\end{figure}

In this paper, we propose a new paradigm for summarizing reviews, rationale-based opinion summarization. Given a set of reviews about an entity (such as a hotel), rationale-based opinion summarization outputs \textit{representative opinions} summarizing the reviews as well as one or more \textit{rationales} for each representative opinion. Fig. \ref{fig:compare_sum} shows an example of a conventional summary produced by a recent extractive summarization model (top) and a rationale-based summary (bottom) containing representative opinions (in blue) and corresponding rationales (in green) for the same entity, a hotel in this case. 
For illustration, we show only one rationale per representative opinion in the figure but in practice, there can be several such rationales specified by users. Such rationale-based summaries can be more useful to users by providing representative opinions as well as informative rationales for them, helping users in making decisions. 

Rationale-based opinion summarization presents several major challenges: (i) what makes a good rationale? and (ii)  how to extract rationales? 
To address the first challenge, we define four desirable properties for rationales: \textbf{relatedness}, \textbf{specificity}, \textbf{popularity}, and \textbf{diversity}. 
 To address the second challenge, we present methods to estimate these properties for review sentences and a Gibbs-sampling-based approach to extract review sentences that can serve as rationales.

\begin{figure}[t]
\centering
\includegraphics[width=0.48\textwidth, keepaspectratio]{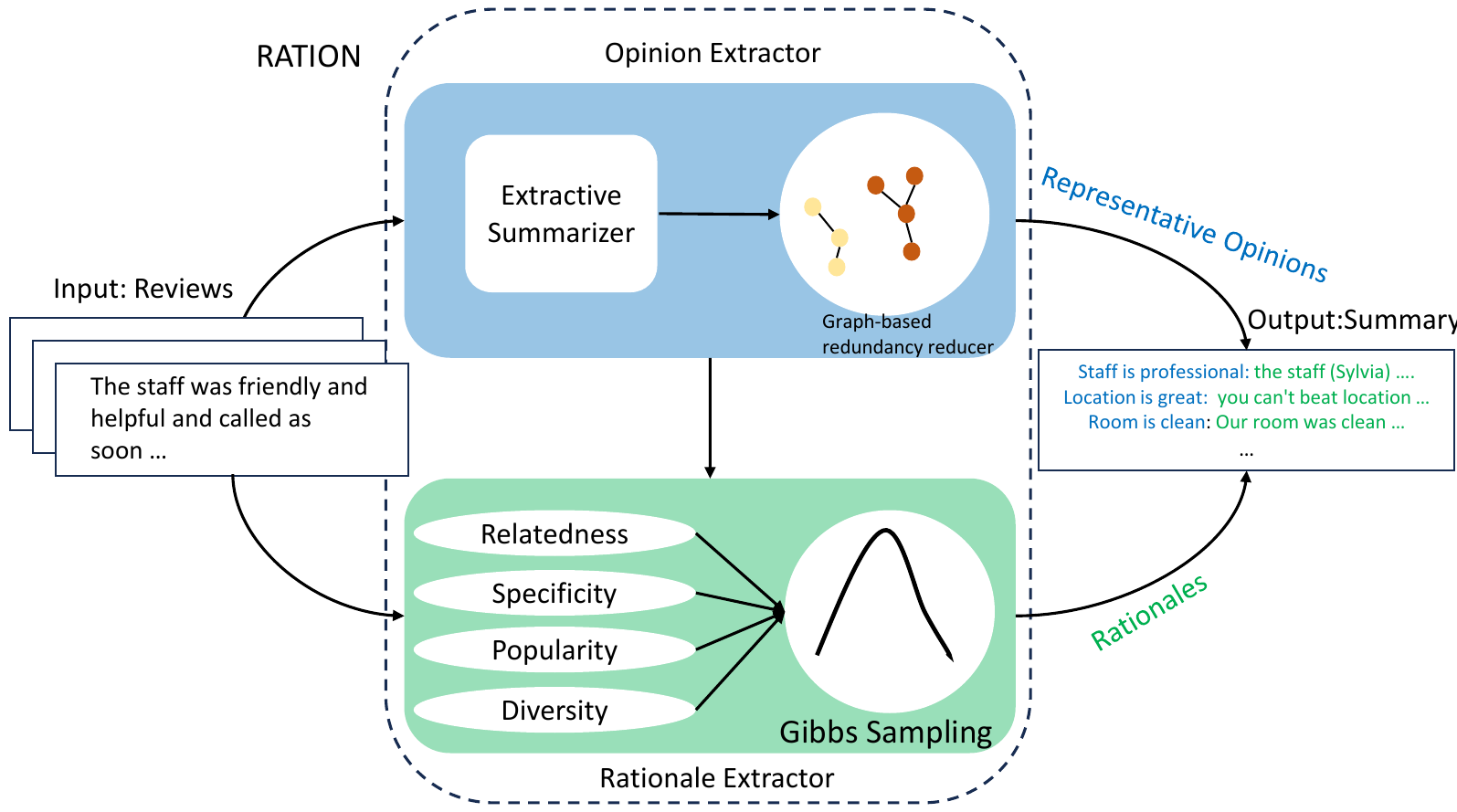}
\caption{Overview of \Ra and its two components: the Opinion Extractor and the Rationales Extractor.} 
\label{fig:compare}
\end{figure}

Overall, we propose \Ra (see Fig.~\ref{fig:compare}), an unsupervised extractive system that has two components: an \textit{Opinion Extractor} (to extract representative opinions) and a \textit{Rationales Extractor} (to extract corresponding rationales). Both the representative opinions and corresponding rationales are extracted from the input review sentences in an unsupervised manner and are presented together as the final output summary. 
The Opinion Extractor extracts representative opinions about various aspects of the entity in a concise manner and removes redundancy in them through a graph-based approach. 
The Rationales Extractor first estimates the four above-mentioned properties of good rationales. 
Since there is no supervision in the review domain for estimating some of these properties, \Ra uses an alignment 
model fine-tuned to the domain of reviews using artificially constructed samples. 
The values of these properties collectively represent the joint probability of a set of review sentence to serve as rationales.
For each representative opinion, \Ra uses Gibbs Sampling to sample a user-specified number of sentences as rationales by approximating this joint probability distribution. 

Our experiments show that rationale-based opinion summaries generated by \Ra are more informative and useful than conventional summaries and the rationales generated by \Ra are better than those generated by strong baselines. 
Our contributions are three-fold:
\begin{itemize}[topsep=1pt, leftmargin=*, noitemsep]
    \itemsep0mm
    \item We propose a new paradigm for summarizing reviews, rationale-based opinion summarization;
    \item We design \Ra, a model to extract representative opinions and corresponding rationales;
    \item We evaluate \Ra using automatic metrics and human evaluation and show that it outperforms strong baselines.
\end{itemize}

\section{Related Work}
There are generally two types of opinion summarization: abstractive 
and extractive.
For abstractive summarization, previous works either use 
aggregate review sentence representations ~\cite{chu2019meansum,isonuma2021unsupervised} or generate synthetic datasets to train generation models in a supervised setting~\cite{bravzinskas2019unsupervised,amplayo-lapata-2020-unsupervised}. For extractive summarization, previous works generally predict the salience of review sentences based on their distance from the aspect representation ~\cite{angelidis2021extractive}, from the average sentence representation \cite{basu-roy-chowdhury-etal-2022-unsupervised} or from the aspect cluster centers \cite{li2023aspect} and extract salient sentences as summaries. However, opinion summaries generated by previous works are usually generic and lack supporting evidence.

To generate more specific opinion summaries, \cite{iso2021convex} generates summaries based on the convex aggregation of review sentence representations instead of the average. However, such summaries might still lack supporting evidence. For explainability, \citet{suhara2020opiniondigest} cluster the opinions extracted from review sentences and generate summaries based on the clusters. \citet{bar-haim-etal-2021-every} matches review sentences to \textit{key points} and extracts the key points that are matched by most review sentences as summaries. \citet{hosking-etal-2023-attributable} generates path representation for each review sentence and generates summaries based on the selected paths. These works can attribute their summary content to a group of review sentences. However, since their goal is to explain the choice of the summary content, the sizes of these groups of supporting sentences are too large to be useful for user consumption. \Ra aims to address this issue by generating rationale-based opinion summarization where each opinion is supported by a small group of rationales.

\section{Problem Statement}
The input in rationale-based opinion summarization is a set of review sentences $S=\{s_1,...,s_n\}$ of a given entity, such as a hotel. The output is a summary $D$ that consists of representative opinions $O=\{o_1,...,o_m\}$ and corresponding sets of rationales $\mathcal{R}=\{R_{1}, R_{2}... R_{m}\}$, where $R_{i} = \{r_{i,1}, r_{i,2}... r_{i,k}\}$, $r_{i,*}$ is a rationale, and $k$ is specified by the user. See Fig.~\ref{fig:compare_sum} for examples of representative opinions and rationales. 

\section{\Ra}

\Ra addresses this problem 
using two components: an \textit{Opinion Extractor} (\S \ref{oe}) and a \textit{Rationales Extractor} (\S \ref{cand}). The representative opinions and rationales are extracted from the input review sentences in an unsupervised manner. They are combined to form a summary, $D$ (\S \ref{summary}). 
\Ra uses an alignment model in its processing which is described in \S \ref{entail}.



\subsection{Opinion Extractor}
\label{oe}
In this section, we describe how \Ra extracts representative opinions $O$ from input review sentences $S$. Representative opinions should be concise sentences that summarizes the reviewers' impressions 
of the entity.
Since existing summarization models are good at identifying this information, \Ra uses an existing extractive opinion summarization model to extract summarizing review sentences. Fig. \ref{fig:compare_sum} (top) shows an example. 

From these summary sentences, \Ra extracts representative opinions of the form `A is B'. For example, from the review sentence, `The hotel was in a great location, fabulous views, and fantastic service.', one representative opinion extracted by \Ra is `location is great'. We chose this format because it is concise yet informative. 
For extracting representative opinions from sentences, \Ra uses a transformer \cite{vaswani2017attention} model proposed by \citet{miao2020snippext} that was finetuned on a ABSA dataset \cite{miao2020snippext, cai-etal-2021-aspect}. 
The ABSA dataset consists of review sentences like `Staff at the hotel is helpful.' annotated with the aspect the sentence is talking about (`service' in this example), sentiment (`positive'), and pairs of nouns and adjectives where adjectives describes the nouns (`staff','helpful'). For a given sentence, \Ra uses the model finetuned on this ABSA dataset to extract pairs of nouns and adjectives. \Ra then concatenates the nouns and adjectives in the form of `noun is adjective' like `staff is helpful' to generate the representative opinions. 
However, since the extracted summarizing sentences are often repetitive, many extracted representative opinions are similar to each other, like `room is spacious' and `room is large'. 
\Ra removes the redundancy among the extracted representative opinions based on their relationship with review sentences. 
It assumes that if two representative opinions are related to a similar group of review sentences, they are likely to be similar. 
For this, it first estimates the relatedness between a representative opinion $o$ and review sentence $s$, using an alignment model $M_{align}$ (described in detail later in \S \ref{entail}). \Ra uses the probability $p_{align}(s,o)$ estimated by $M_{align}$ that $s$ aligns with $o$ as the relatedness. 
Next, using this relatedness, \Ra estimates the similarity between two representative opinions $o$ and $o'$. For this, it constructs a feature vector for every representative opinion, $o$, $f_o\in R^n$ whose $i$-th element is $p_{align}(s_i,o)$ if review sentence $s_i$ aligns with $o$, otherwise it is zero. The similarity between two representative opinions $o$ and $o'$, is defined as the cosine similarity between their feature vectors $f_o$ and $f_{o'}$. Next, to cluster similar representative opinions together, \Ra constructs an undirected graph where each node is a representative opinion and there is an edge between two nodes if their similarity is greater than a threshold $\beta$.  
Each connected component of the graph forms an \textit{opinion cluster} $G$ and its most prototypical node (the node that is aligns with the most review sentences) is extracted as a representative opinion $o_i \in O$. The number of representative opinions in $O$ is equal to the number of clusters identified above. 

\subsection{Rationales Extractor}
 \label{cand}
In this section, we describe how for each representative opinion $o_i$, \Ra extracts a set of $k$ rationales, $R_i$, from the input review sentences $S$. 

For a given representative opinion, $o_i$, not all review sentences are viable candidates for its rationales since they might not be relevant to it. 
We filter out such nonviable candidates and retain only viable ones as the \textit{rationale candidate set} $C_i$ using the alignment model, $M_{align}$. 
Let $G_i$ represent the opinion cluster that representative opinion $o_i$ belongs to. A review sentence, $s$, is included in the candidate set $C_i$ if (i) it aligns with at least one opinion in $G_i$, and (ii) it is most related to $G_i$ among all clusters. \Ra defines the relatedness between review sentence $s$ and cluster $G$ as the maximum alignment score between $s$ and any element of $G$: 
\begin{equation}
e(s,G)=max_{o \in G} p_{ent}(s,o)
\label{eqn:sent_clus_rel}
\end{equation}



After removing nonviable candidates, \Ra extracts rationales, $R_i$,  from the rationale candidate set, $C_i$,  for each representative opinion $o_i$. 
Good rationales should be related to the corresponding representative opinion (relatedness). They should contain specific details (specificity), represent popular information (popularity), and offer diverse information (diversity). We now describe how to quantify these properties and then describe how to extract rationales based on these properties.

\noindent\textbf{Relatedness} of review sentence $s$ to representative opinion $o_i$, ($rel(s)$), measures how related $s$ is to $o_i$ as compared to all other representative opinions. As before, let $G_i$ represent the cluster that $o_i$ belongs to. Using the definition of relatedness between a review sentence $s$ and a cluster $G$ (Equation~\ref{eqn:sent_clus_rel}), $rel(s)$ is defined as:
\vspace{-0.2cm}
  \begin{equation}
  rel(s)=\frac{e(s,G_i)}{\sum_{G_k \in G_s} e(s,G_k)}
 \end{equation}
\noindent where $G_s$ is the set of the opinion clusters that has at least one element that sentence $s$ aligns with.

\noindent\textbf{Specificity} of review sentence $s$, ($spec(s)$), measures the amount of details that $s$ contains. For this, it uses a Deberta \cite{hedeberta} model finetuned on a specificity estimation dataset \cite{ko2019domain}. 

\noindent\textbf{Popularity} of review sentence $s$, ($pop(s)$), measures how representative it is of the rationale candidate set it belongs to. 
 To calculate $pop(s)$, \Ra constructs a weighted undirected graph. The nodes of this graph represent the review sentences in the rationale candidate set $C_i$, $s\in C_i$. The representative opinion $o_i$ also forms a node. There is an edge between two review sentences if one aligns with the other or vice versa (as estimated by $M_{align}$). The weight of this edge is the greater of the two alignment probabilities. There is an edge between a review sentence and the representative opinion if the review sentence aligns with the representative opinion and the weight of this edge is the alignment probability. 
 \Ra measures the popularity $pop(s)$ of sentence $s$ as the centrality of the corresponding node in this graph. 
 
\noindent\textbf{Diversity} of a group of review sentences, $s_{1:k}$, ($div(s_{1:k})$), measures how dissimilar their content collectively is. 
It is estimated as the negative of the pairwise cosine similarity of their bag-of-word representations.

\noindent\textbf{Gibbs Rationale Sampler:} Based on the properties defined above, \Ra defines the joint probability of a group of review sentences, $s_{1:k}$, to be selected as rationales to be proportional to:
\vspace{-0.2cm}
\begin{equation}
 exp(\sum_{i=1}^k sal(s_{i})+\gamma div(s_{1:k}))
 \label{eq:prob}
 \end{equation}
\noindent where $\gamma>0$ is the weight of the diversity term and $sal(s)$ is the product of $rel(s)$, $spec(s)$ and $pop(s)$, each normalized to $[0,1]$ using min-max normalization among the rationale candidate set $s$ belongs to.

However, directly computing this probability for all possible groups is computationally expensive. 
To address this issue, \Ra uses Gibbs Sampling. 
Gibbs Sampling is a  Markov chain Monte Carlo algorithm that can approximate the joint probability of a group of sentences $s_{1:k} \subset C_i$ being considered as rationales, $R_{i} =\{r_{i1}, r_{i2} ... r_{ik}\}$, for the representative opinion, $o_i$. 
Since the joint probability is difficult to sample from, it iteratively samples individual $r_{i*}$ conditioned on the values of other $r_{i*}$s. The sequence of samples hence obtained form a Markov chain and its stationary distribution approximates the joint distribution. 
Using $R_{i\neg j}$ to refer to all elements of $R_{i}$ except the $j^{th}$ element $r_{ij}$, the conditional probability $p(r_{ij}=s^*|R_{i\neg j})$ is proportional to:
\vspace{-0.2cm}
 \begin{equation}
 \frac{exp(sal(s^*) + \gamma div(\{R_{i\neg j}, s^*\}))}{\sum_{s \in C_i} exp(sal(s) + \gamma div(\{R_{i\neg j}, s\}))}
 \end{equation}

This sampling process is detailed in Alg. \ref{alg:1}. The input of the algorithm is the representative opinion $o_i$, its rationale candidate set $C_i$,and $\eta,\theta$ (Line \ref{input}). Initially, $R_i$ are randomly sampled from rationale candidate set $C_i$ (Line \ref{init}). In each Gibbs update, 
$r_{i\cdot}$ 
is sampled from the conditional distribution conditioned on other sentences, $R_{i\neg j}$(Line \ref{cond_prob}). 
After the \textit{burn-in period} of $\eta$, \Ra records the frequency of sampled review sentence group in additional $\theta$ scans as $R_i$ to approach the stationary distribution more closely 
(Line \ref{record}). \Ra extracts the most frequent review sentence group as the rationales $R_i$ (Line \ref{frequent}). We show example rationale candidate sets and extracted rationales in Figure \ref{fig:sample_rationale_candidate1} and Figure \ref{fig:sample_rationale_candidate2}. 

\begin{algorithm}[t!]
\caption{Gibbs Rationale Sampler}
\footnotesize
\begin{algorithmic}[1]
\State \textbf{Input}: $\eta$, $\theta$, $o_i$, $C_i$ \label{input}
\State Randomly initialize $R_i$ from $C_i$ \label{init}
\State R=\{\} \Comment{R records the frequency of sentence groups}
\For{$l=1$ to $\eta+\theta$}
\For{$j=1$ to $k$}
\State sample $r_{ij}\sim p(r_{ij}=s^*|R_{i\neg j})$\label{cond_prob}
\If{$l>\eta$}
\State R[$R_i$]+=1  \EndIf \label{record}
\EndFor
\EndFor
\State $R_i$=$\argmax_R'$R[$R'$] \label{frequent}
\State \textbf{return} $R_i$
\end{algorithmic}
\label{alg:1}
\end{algorithm}

\subsection{Summarization}
\label{summary}
We now describe how \Ra generates summary $D$ using representative opinions $O$ and rationales $\mathcal{R}$. 
In principle, \Ra can simply pair each $o_i \in O$ with the rationales in corresponding $R_i \in \mathcal{R}$. 
However, sometimes the user might want to put restrictions on the length of the summary.
In such cases, \Ra gives more importance to representative opinions supported by more review sentences. 
It obtains them by ranking the representative opinions in $O$ in descending order of the size of the corresponding rationale candidate sets.
\Ra then constructs the summary, $D$, by picking representative opinions $o_i$ from this ranked list and the corresponding rationales $R_i$ until the length limit is reached (examples shown in Appendix Fig. \ref{fig:sample}). 

\subsection{The Alignment Model}
\label{entail}
At various stages in its processing, \Ra uses an alignment model $M_{align}$ to estimate alignment or relatedness between pairs of sentences.  $M_{align}$ takes a pair of sentences $\langle$X, Y$\rangle$ as input, and predicts whether X aligns with Y (\textit{alignment}), X opposes Y (\textit{opposite}) or X is neutral to Y (\textit{neutral}). However, there is no in-domain supervision available for finetuning this alignment model. \Ra therefore finetunes a RoBerta \cite{radford2019language} model on artificially generated samples from the ABSA dataset (described in \S \ref{oe}). 
It generates two types of fine-tuning samples: \textit{Sent-Opinion} pairs and \textit{Sent-Sent} pairs. 




\noindent\textbf{Sent-Opinion Pairs:} \Ra uses $M_{align}$ to estimate alignment between review sentences and representative opinions (\S \ref{oe}). 
To enable this learning, we construct \textit{alignment} samples for fine-tuning $M_{align}$ by pairing a sentence, $s$, from the ABSA dataset (X) with the representative opinion extracted from itself (Y) using the method described in \S \ref{oe}. 
For \textit{neutral} pairs, the second sentence, Y, is a representative opinion obtained from other sentences that have the same sentiment as $s$ but discuss a different category. 
For \textit{opposite} pairs, the second sentence, Y, is a representative opinion obtained from other sentences with the same category as $s$ but an opposite sentiment. 

\noindent\textbf{Sent-Sent Pairs:} \Ra also uses $M_{align}$ to estimate alignment between review sentences (\S \ref{cand}). 
To enable this learning, we construct \textit{alignment} samples as before for neutral pairs and opposite pairs except that instead of pairing sentences (X) with representative opinions extracted from randomly sampled sentences, we pair them with the sampled sentences themselves (Y). For alignment pairs, the second sentence Y are a randomly sampled sentence with the same aspect and sentiment as X.

\section{Empirical Evaluation}
We now describe experiments to evaluate \Ra.
\subsection{Implementation Detail} 
For the Opinion Extractor, \Ra uses SemAE \cite{basu-roy-chowdhury-etal-2022-unsupervised} as the extractive summarization model but our method is independent of this choice. We only assume the existence of extractive summaries. We also perform experiments on the extractive summaries generated by $\rm Hercules$ \cite{hosking-etal-2023-attributable} (Appendix \ref{sec:hercules_experiment}). 
From the summarizing review sentences,  \Ra uses Snippext \cite{miao2020snippext} as the ABSA model to extract representative opinions.

For the Rationales Extractor, to accelerate the calculation of the alignment probability, $p_{align}$, we use a sentiment classification model \cite{barbieri-etal-2020-tweeteval}. Specifically, when the two input sentences do not have the same sentiment label, we directly set their $p_{align}$ to $0$. 
When extracting rationales, we extract clauses instead of full sentences since we find clauses are more specific to representative opinions than full sentences. We describe the process of dividing sentences into clauses in the appendix \ref{segment}. 
We also filter out rationale candidate set $C$ with less than five sentences. 
When estimating popularity $pop(s)$, we use the default TextRank for an undirected graph to estimate the centrality of the node. 
For estimating $spec(s)$, we finetune a DeBERTa-base \cite{hedeberta} model on the specificity dataset for 3 epochs with the learning rate as 2e-5 and batch size as 32. The weight of the diversity term $\gamma$ is 0.1. As for Gibbs Sampling, $\eta$ is 100 and $\theta$ is 200. When sampling from the conditional probability, we set the temperature of Softmax as 0.01.

For the alignment model $M_{align}$, we use one alignment models for the Space data and the Yelp dataset respectively. We first perform domain adaptation using sentences sampled from the corresponding train sets 
on RoBERTa-large \cite{liu2019roberta} following steps described in \citet{bar-haim-etal-2021-every}. To generate in-domain pairs to finetune the alignment model $M_{align}$,  aside from sentences in the corresponding ABSA dataset, we additionally sample sentences from the corresponding train set to create a dataset containing 7,000 sentences. The annotations of the sampled sentences are predicted by the same ABSA model that \Ra uses for the Opinion Extractor. For each sentence, we generate one Sent-Opinion pairs and Sent-Sent pairs for each label.
We then perform down-sampling to create a dataset containing 24K samples for the Space data and the Yelp dataset respectively and use about 20K of them for training. We use the remaining samples for validation. 
The size of the dataset matches the size of ArgKP dataset \cite{bar-haim-etal-2020-arguments} for the fair comparison we described in \S \ref{sec:ent_eval}. 
We then finetune $M_{align}$ on the in-domain datasets for 3 epochs with the learning rate as 1e-5 and batch size as 32. 

\subsection{Dataset}

We perform the experiments on the Space dataset \cite{angelidis2021extractive} and the Yelp dataset\footnote{https://www.yelp.com/dataset}. 
For the Space dataset, we held out randomly sampled $250$ entities with $100$ reviews each as the test set. The remaining data was used for training and development. 
For the Yelp dataset, we perform cleaning and downsampling (Appendix \ref{sec:prep_yelp}) 
and only retain entities whose categories contain `restaurant'. From these entities, we sample $50$ and $250$ entities with $100$ reviews each as the development set and the test set respectively. The statistics of datasets are shown in Appendix Table \ref{tab:dataset_stat}. 
We tune the hyperparameters on the development sets and report the performance on the test sets. 

To finetune the ABSA model used in Opinion Extractor and produce fine-tuning samples for $M_{align}$, we use the ABSA dataset in the hotel domain 
for the Space dataset, and ACOS-restaurant dataset~\cite{cai-etal-2021-aspect} for the Yelp dataset.


\subsection{Rationale-based Summary Evaluation}
\label{sec:summary}
We compare rationale-based opinion summaries generated by \Ra with conventional summaries generated by a state-of-the-art opinion summarization model, SemAE \cite{basu-roy-chowdhury-etal-2022-unsupervised} using human evaluation. We ask annotators to compare the two types of summaries in a pairwise manner based on four criteria: which summary includes more information (\textit{informativeness}), which summary contains less repeated phrases (\textit{non-redundancy}), which summary is easier to read (\textit{coherence}), and which summary is more useful for decision making (\textit{usefulness}). We randomly sample 25 entities each from the test sets of the Space dataset and the Yelp dataset and generate 100-word summaries for each entity using \Ra and SemAE. Each pair of summaries is annotated by three annotators recruited from Amazon Mechanical Turk (AMT). The human annotators are required to be in the United States, have HIT Approval Rate greater than 98, and be AMT masters. Fig.~\ref{fig:human_eval2} in the Appendix shows a screenshot of our setup. 
In Table \ref{tab:summ_comp}, we report the win rates of \Ra, the win rates of SemAE, and the tie rates between the two on the four criteria: informativeness, non-redundancy, coherence, and usefulness.

\begin{table}[]
\centering
\small
\begin{tabular}{lcccc}
\hline
       & Info. & Non-Redun.    & Cohe.         & Use.          \\ \hline
       & \multicolumn{4}{c}{Space}                             \\
\Ra & 0.52  & \textbf{0.88} & \textbf{0.84} & \textbf{0.68} \\
SemAE  & 0.32  & 0.04          & 0.12          & 0.28          \\
Tie    & 0.16  & 0.08          & 0.04          & 0.04          \\ \hline
       & \multicolumn{4}{c}{Yelp}                              \\
\Ra & 0.36  & \textbf{1.00} & \textbf{0.68} & \textbf{0.56} \\
SemAE  & 0.36  & 0.00          & 0.04          & 0.2           \\
Tie    & 0.28  & 0.00          & 0.28          & 0.24 \\ \hline         
\end{tabular}
\caption{Human comparison of rationale-based opinion summaries generated by \Ra with conventional summaries generated by SemAE. \textbf{Bold} numbers indicate significant differences ($p$<0.05, paired bootstrap resampling \cite{koehn-2004-statistical}). Rationale-based opinion summaries outperform conventional opinion summaries on non-redundancy, coherence, and usefulness. }
\label{tab:summ_comp}
\end{table}

From the table, we can see that rationale-based summaries perform significantly better on non-redundancy, coherence, and usefulness 
than conventional summaries. Rationale-based summaries do not perform very well on informativeness because they pair each representative opinion with rationales. 
 Therefore, while they provide more information per representative opinion, they understandably do not cover all opinions expressed in the conventional summaries because of the length limit. This can easily be fixed by increasing the length limit. We also perform error analysis of the summaries generated \Ra (Appendix \ref{sec:error}). Overall, the experiments indicate that rationale-based summaries are less redundant, easier to read, and more useful for decision-making. 

\subsection{Rationale Evaluation}
\label{sec:ration}
We evaluate the extracted rationales using automatic (\S \ref{auto}) and human measures (\S \ref{human}).
\subsubsection{Automatic Evaluation}
\label{auto}
We use the following four automatic measures for evaluating rationales for a given opinion.

To measure \textbf{relatedness} between the rationales and the corresponding representative opinion, $emb_{rel}$, we use the average cosine similarity between the sentence embeddings (obtained using SimCSE \cite{gao2021simcse}) of the representative opinion and each of its rationales.

To measure \textbf{specificity}, $key_{spec}$, we use TF-IDF-based keywords. For this, we concatenate all review sentences belonging to the same rationale candidate set and calculate TF-IDF scores based on the concatenated sentences from each rationale candidate set of an entity. For each rationale candidate set, we extract five words with the highest TF-IDF scores that are not part of the representative opinions as the keywords. These keywords represent the popular details about the representative opinion but are not directly present in it. Given a set of rationales, $key_{spec}$ is the sum of TF-IDF scores of the keywords covered by that set divided by the sum of TF-IDF scores of all keywords. 

To measure \textbf{popularity}, $key_{pop}$, we consider the fraction of rationales' tokens that are keywords. Given a set of rationales, $key_{pop}$ is the sum of TF-IDF scores of the keywords covered by them divided by the sum of TF-IDF scores of all tokens present in the rationales. 

To measure \textbf{diversity} among the rationales, ($emb_{div}$), we use one minus the average pairwise cosine similarity of their sentence embeddings.

Based on these four measures, we compare \Ra with its variants: \Ra (w/o X). \Ra (w/o X) represents a variant of \Ra that does not consider X for the probability of being rationales (Eqn. \ref{eq:prob}). We also compare \Ra with InstructGPT~\cite{ouyang2022training} version `gpt-3.5-turbo-0613', and two extractive opinion summarization models: SemAE~\cite{basu-roy-chowdhury-etal-2022-unsupervised} and TokenCluster~\cite{li2023aspect}. 
To extract rationales using InstructGPT, we provide a representative opinion and its corresponding rationale candidate set as input and prompt IntructGPT to extract a predefined number of rationales from the rationale candidate set. The prompt also describes the four desirable properties of rationales (shown in Appendix \ref{sec:instruct}). 
To extract rationales using SemAE and TokenCluster, we provide the rationale candidate set of a representative opinion as input and use these models to extract a conventional opinion summary of the candidates set with a predefined number of sentences. We treat each sentence of the summary as a rationale for the representative opinion. We use these models as baselines because they can extract popular information from the input. However, they do not consider other aspects of rationale extraction, so the comparison is unfair to them. Therefore, we treat InstructGPT as the primary baseline in this experiment.
We evaluate in two different settings: k=1 and k=3, where k is the number of rationales extracted for each representative opinion. The results are shown in Table \ref{tab:auto}. In addition to the four measures, we also report an \textit{Overall} score which is the average of normalized values ($[0,1]$) of these measures. 


\begin{table}[t]
\centering 
\resizebox{0.47\textwidth}{!}{
\begin{tabular}{lccccc}
\hline
            & $emb_{rel}$ & $key_{spec}$ & $key_{pop}$ & $emb_{div}$ & Overall \\ \hline
            & \multicolumn{5}{c}{Space (k=1)}             \\
\Ra          &   0.422      & 0.217     & 0.224    &    -    & 0.710  \\
~~~w/o $rel$      & 0.423        & 0.223     & 0.225    &  -    & \textbf{0.740}    \\
~~~w/o $spec$     & 0.555        & 0.147     & 0.208    &   -   & 0.559   \\
~~~w/o $pop$       & 0.369         & 0.195     & 0.193    &  -   &  0.445   \\
InstructGPT &  0.586        & 0.139     & 0.129    &   -    &  0.294 \\ 
SemAE & 0.615 & 0.165 & 0.191 & -  & 0.650 \\
TokenCluster & 0.610 & 0.142 & 0.177 & - & 0.505 \\
\hdashline
            & \multicolumn{5}{c}{Space (k=3)}             \\
\Ra          &   0.414       & 0.498     & 0.224    &  0.580   &    \textbf{0.730} \\
~~~w/o $rel$       & 0.415         & 0.499     & 0.223    &  0.575   &  0.724   \\
~~~w/o $spec$       &  0.508        & 0.420     & 0.222    & 0.528   & 0.625     \\
~~~w/o $pop$        &  0.377        & 0.465     & 0.203    &  0.627  &  0.508    \\
~~~w/o $div$       &   0.418       & 0.487     & 0.226    &   0.564   & 0.716    \\ 
InstructGPT        &   0.501       & 0.438 &   0.193 
     & 0.530  &  0.434 \\ 
SemAE & 0.593 & 0.374 &  0.215 & 0.349 & 0.413 \\
TokenCluster & 0.563 & 0.400 & 0.204 & 0.424 & 0.417 \\ \hline
            & \multicolumn{5}{c}{Yelp (k=1)}              \\
\Ra          &  0.358        & 0.202     & 0.233    &  -    &  \textbf{0.718}  \\
~~~w/o $rel$        & 0.372         & 0.201     & 0.226    &  -   & 0.713    \\
~~~w/o $spec$      &  0.469        & 0.154     & 0.208    & -   &  0.626    \\
~~~w/o $pop$      &  0.313        & 0.180     & 0.198    &  -   & 0.504    \\
InstructGPT     &   0.604       & 0.095      & 0.111    &  -  &  0.333    \\ 
SemAE & 0.599 & 0.116 & 0.148 & - & 0.497 \\ 
TokenCluster & 0.581 & 0.128 & 0.159 & - & 0.541 \\ \hdashline
            & \multicolumn{5}{c}{Yelp (k=3)}              \\
\Ra          &  0.337        & 0.473     & 0.227    &   0.619    & \textbf{0.734}  \\
~~~w/o $rel$       & 0.347         & 0.472     & 0.225    & 0.601    &  0.710   \\
~~~w/o $spec$      &  0.408        & 0.435     & 0.226    & 0.609    & 0.732     \\
~~~w/o $pop$       &  0.323        & 0.449     & 0.203    &  0.641   & 0.588    \\
~~~w/o $div$       &  0.345       & 0.455     & 0.228    &  0.596   &  0.691  \\ 
InstructGPT &  0.444        & 0.396     & 0.194    &   0.597    &  0.537 \\ 
SemAE & 0.544 & 0.331 & 0.176 & 0.430 & 0.250 \\ 
TokenCluster & 0.508 & 0.370 & 0.182 & 0.488 & 0.380 \\ \hline
\end{tabular}}
\caption{Automatic evaluation of rationales on Space and Yelp datasets with one (k=1) and three (k=3) rationales extracted per representative opinion. Considering the four measures and their \textit{overall} values, \Ra extracts the best rationales. }
\label{tab:auto}
\end{table}

From the table, we can observe that in general, rationales generated by \Ra outperform rationales generated by its variants considering the overall quality. This indicates that all terms in the probablity function (Eqn. \ref{eq:prob}) are important for extracting good rationales. For InstructGPT, we observe that although the instructions ask it to extract rationales with lots of details, some of the extracted rationales are paraphrases of the representative opinions, which is indicated by high $emb_{rel}$ but poor $key_{spec}$. We provide more discussion of rationales generated by InstructGPT in Appendix \ref{sec:instruct_rationale}.

\subsubsection{Human Evaluation}
\label{human}
We also conduct a human evaluation of the rationales generated by \Ra and InstrutGPT \cite{ouyang2022training} for a given representative opinion. We randomly sample $50$ representative opinions each from the entities belonging to the test sets of the Space dataset and the Yelp dataset and generate three rationales for each representative opinion using \Ra and InstructGPT. Each pair of rationale sets is evaluated by three annotators recruited from Amazon Mechanical Turk. The annotator details are same as in Sec.~\ref{sec:summary}.
We ask annotators to compare the two rationale sets in a pairwise manner based on three properties: relatedness, specificity, and diversity. Fig.~\ref{fig:human_eval} of the Appendix shows our setup. In Table \ref{tab:ration_comp}, we report the win rate of \Ra, the win rate of InstructGPT, and the tie rate between the two on the three properties.

\begin{table}[]
\centering
\begin{tabular}{lccc}
\hline
            & Rel. & Spec.         & Div.          \\ \hline
            & \multicolumn{3}{c}{Space}            \\
\Ra      & 0.34 & 0.40          & \textbf{0.42} \\
InstructGPT & 0.34 & 0.30          & 0.22          \\
Tie         & 0.32 & 0.30          & 0.36          \\ \hline
            & \multicolumn{3}{c}{Yelp}             \\
\Ra     & 0.26 & \textbf{0.56} & \textbf{0.64} \\
InstructGPT & 0.38 & 0.08          & 0.10          \\
Tie         & 0.36 & 0.36          & 0.26          \\ \hline
\end{tabular}
\caption{Human evaluation of rationales generated by \Ra and InstructGPT. 
\textbf{Bold} indicates significant differences ($p$<0.05, paired bootstrap resampling). \Ra outperforms InstructGPT on specificity and diversity and is comparable to it on relatedness}. 
\label{tab:ration_comp}
\end{table}
We can see from the table that \Ra outperforms InstructGPT on specificity and diversity. 
For relatedness, both systems were judged to be comparable. Since most review sentences belonging to the rationale candidate set are already quite related to the representative opinion, it is understandable that both systems do not have much difference in relatedness. Overall, the experiments indicate that rationales extracted by \Ra are more specific and diverse than InstructGPT.

\subsection{Rationale Candidate Set Evaluation}
\label{sec:ent_eval}

\begin{table}[t!]
\centering
\resizebox{0.48\textwidth}{!}{
\begin{tabular}{lcccc}
\hline
         & \textit{Silh}     & \textit{NPMI}           & \textit{SC}           & Overall  \\ \hline
         & \multicolumn{4}{c}{Space}                           \\
$\rm RoBERta_{mnli}$   & 0.089 & -0.061 & 0.970 & 0.779 \\
$\rm KPA$      & 0.134 & -0.059 & 0.962 & 0.836 \\
$\rm Snippext$     & 0.108 & -0.103 & 0.934 & 0.265 \\
$\rm Hercules$ & 0.009 & -0.042 & 0.943 & 0.415 \\
\Ra   & 0.119 & -0.051 & 0.969 & \textbf{0.906} \\ \hline
         & \multicolumn{4}{c}{Yelp}                            \\ 
$\rm RoBERta_{mnli}$     & 0.015 & -0.210 & 0.956 & 0.530 \\
$\rm KPA$      & 0.035 & -0.208 & 0.934 & 0.758 \\
$\rm Snippext$     & 0.039 & -0.265 & 0.805 & 0.318 \\
\Ra   & 0.040 & -0.171 & 0.934 & \textbf{0.953} \\ \hline
\end{tabular}}
\caption{Automatic evaluation of rationale candidate sets. Considering the three measures and their overall scores, \Ra generates rationale candidates of better quality than the baselines. }
\label{tab:seteval}
\end{table}


\Ra extracts rationales for a representative opinion from a rationale candidate set instead of all review sentences. In this experiment, we evaluate the goodness of this set by comparing it with adaptations of previous works that match a group of review sentences to summary sentences. 

For this evaluation, we use three automatic measures. First, we view each rationale candidate set as a cluster of sentences and evaluate the clustering quality. We report Silhouettes scores \cite{rousseeuw1987silhouettes} (\textit{Silh}) based on the cosine similarity of the sentence embeddings. Second, we borrow measures from topic modeling to compute coherence of the sets using TF-IDF scores of tokens for coherence. Third, we also report the entailment score (\textit{SC}) between the concatenation of all candidates in a rationale candidate set and the corresponding representative opinion as predicted by SummaC \cite{laban2022summac}. 



We compare \Ra with four baseline models: $\rm RoBERta_{mnli}$, $\rm KPA$, $\rm Snippext$, and $\rm Hercules$. $\rm RoBERta_{mnli}$ \cite {louis-maynez-2023-opinesum} uses a RoBERta-large \cite{liu2019roberta} finetuned on the MNLI dataset \cite{N18-1101} to match reviews to `propositions'. $\rm KPA$ \cite{bar-haim-etal-2021-every} uses a domain adapted RoBERta-large that is then finetuned on ArgKP dataset \cite{bar-haim-etal-2020-arguments} to match review sentences to `key points'. $\rm Snippext$ \cite{miao2020snippext} is trained on ABSA datasets to estimate the aspect and the sentiment distributions for each opinion and then uses similarity between these distributions to cluster `opinions'.  $\rm Hercules$ \cite{hosking-etal-2023-attributable} generates a path on a tree for each review sentence and summary and then uses path similarity to match review sentences to summary sentences. We use these models to estimate alignment between representative opinions and review sentences (details in Appendix \ref{sec:detail_candidate}), and generate rationale candidate sets accordingly as in \S \ref{cand}. 
Because of the different ranges of these measures, we normalize these measures to $[0,1]$ among all baselines and use the average of these normalized metrics to evaluate the overall quality of rationale candidate sets. 
The results are shown in Table \ref{tab:seteval}. In addition to the three measures, we also report an \textit{Overall} score which is the average of normalized values ($[0,1]$) of these measures. 

From the table, we can observe that the rationale candidate sets generated by \Ra have the best overall performance. The result shows the effectiveness of the in-domain pairs we created to finetune the alignment model.





\section{Conclusion}
We propose rationale-based opinion summarization, a new paradigm for summarizing reviews. The rationale-based summaries present representative opinions and their corresponding rationales. We define four desirable properties of rationales: relatedness, specificity, popularity, and diversity. Based on these properties, we propose \Ra, an unsupervised extractive system that extracts representative opinions and their corresponding rationales based on Gibbs sampling. Our experiments show that rationale-based summaries generated by \Ra are more useful than conventional opinion summaries. Our experiments also show that the rationales generated by \Ra outperform its variants and strong baselines. 

\section{Limitation}
\label{sec:limit}
Since there is no supervision for extracting rationales, \Ra separately estimates the four properties of rationales separately and assign equal importance to relatedness, specificity, and popularity. Future work can collect supervised data to extract rationales and build a system that can jointly model and assign weights to the four properties based on the supervised data. Second limitation is during extracting rationales, \Ra does not consider the similarity between representative opinions. Another limitation is that our datasets and all experiments are only focused on the English language. 

\section{Ethical Consideration}
We do not expect any ethical risks caused by our work. The datasets we use are all publicly available. We do not annotate any data on our own. 
We performed human evaluation experiments on Amazon Mechanical Turk. The annotators were compensated at a rate of \$15 per hour. During the evaluation, human annotators were not exposed to any sensitive or explicit content.

\section{Acknowledgment}
The authors appreciate the helpful feedback from Mohit Bansal, Somnath Basu Roy Chowdhury, Anvesh Rao Vijjini, and Anneliese Brei in the early phase of this work. This work was partly supported by the National Science Foundation under award DRL-2112635. 

\bibliography{emnlp2023}

\begin{thebibliography}{33}
\expandafter\ifx\csname natexlab\endcsname\relax\def\natexlab#1{#1}\fi

\bibitem[{Amplayo and Lapata(2020)}]{amplayo-lapata-2020-unsupervised}
Reinald~Kim Amplayo and Mirella Lapata. 2020.
\newblock Unsupervised opinion summarization with noising and denoising.
\newblock In \emph{Proceedings of the 58th Annual Meeting of the Association for Computational Linguistics}, pages 1934--1945, Online. Association for Computational Linguistics.

\bibitem[{Angelidis et~al.(2021)Angelidis, Amplayo, Suhara, Wang, and Lapata}]{angelidis2021extractive}
Stefanos Angelidis, Reinald~Kim Amplayo, Yoshihiko Suhara, Xiaolan Wang, and Mirella Lapata. 2021.
\newblock Extractive opinion summarization in quantized transformer spaces.
\newblock \emph{Transactions of the Association for Computational Linguistics}, 9:277--293.

\bibitem[{Bar-Haim et~al.(2020)Bar-Haim, Eden, Friedman, Kantor, Lahav, and Slonim}]{bar-haim-etal-2020-arguments}
Roy Bar-Haim, Lilach Eden, Roni Friedman, Yoav Kantor, Dan Lahav, and Noam Slonim. 2020.
\newblock From arguments to key points: {T}owards automatic argument summarization.
\newblock In \emph{Proceedings of the 58th Annual Meeting of the Association for Computational Linguistics}, pages 4029--4039, Online. Association for Computational Linguistics.

\bibitem[{Bar-Haim et~al.(2021)Bar-Haim, Eden, Kantor, Friedman, and Slonim}]{bar-haim-etal-2021-every}
Roy Bar-Haim, Lilach Eden, Yoav Kantor, Roni Friedman, and Noam Slonim. 2021.
\newblock Every bite is an experience: {K}ey {P}oint {A}nalysis of business reviews.
\newblock In \emph{Proceedings of the 59th Annual Meeting of the Association for Computational Linguistics and the 11th International Joint Conference on Natural Language Processing (Volume 1: Long Papers)}, pages 3376--3386, Online. Association for Computational Linguistics.

\bibitem[{Barbieri et~al.(2020)Barbieri, Camacho-Collados, Espinosa~Anke, and Neves}]{barbieri-etal-2020-tweeteval}
Francesco Barbieri, Jose Camacho-Collados, Luis Espinosa~Anke, and Leonardo Neves. 2020.
\newblock {T}weet{E}val: Unified benchmark and comparative evaluation for tweet classification.
\newblock In \emph{Findings of the Association for Computational Linguistics: EMNLP 2020}, pages 1644--1650, Online. Association for Computational Linguistics.

\bibitem[{Basu Roy~Chowdhury et~al.(2023)Basu Roy~Chowdhury, Monath, Dubey, Ahmed, and Chaturvedi}]{basu-roy-chowdhury-etal-2023-unsupervised-opinion}
Somnath Basu Roy~Chowdhury, Nicholas Monath, Kumar Dubey, Amr Ahmed, and Snigdha Chaturvedi. 2023.
\newblock Unsupervised opinion summarization using approximate geodesics.
\newblock In \emph{Findings of the Association for Computational Linguistics: EMNLP 2023}, pages 97--112, Singapore. Association for Computational Linguistics.

\bibitem[{Basu Roy~Chowdhury et~al.(2022)Basu Roy~Chowdhury, Zhao, and Chaturvedi}]{basu-roy-chowdhury-etal-2022-unsupervised}
Somnath Basu Roy~Chowdhury, Chao Zhao, and Snigdha Chaturvedi. 2022.
\newblock Unsupervised extractive opinion summarization using sparse coding.
\newblock In \emph{Proceedings of the 60th Annual Meeting of the Association for Computational Linguistics (Volume 1: Long Papers)}, pages 1209--1225, Dublin, Ireland. Association for Computational Linguistics.

\bibitem[{Bird et~al.(2009)Bird, Klein, and Loper}]{bird2009natural}
Steven Bird, Ewan Klein, and Edward Loper. 2009.
\newblock \emph{Natural language processing with Python: analyzing text with the natural language toolkit}.
\newblock " O'Reilly Media, Inc.".

\bibitem[{Bra{\v{z}}inskas et~al.(2019)Bra{\v{z}}inskas, Lapata, and Titov}]{bravzinskas2019unsupervised}
Arthur Bra{\v{z}}inskas, Mirella Lapata, and Ivan Titov. 2019.
\newblock Unsupervised opinion summarization as copycat-review generation.
\newblock \emph{arXiv preprint arXiv:1911.02247}.

\bibitem[{Cai et~al.(2021)Cai, Xia, and Yu}]{cai-etal-2021-aspect}
Hongjie Cai, Rui Xia, and Jianfei Yu. 2021.
\newblock Aspect-category-opinion-sentiment quadruple extraction with implicit aspects and opinions.
\newblock In \emph{Proceedings of the 59th Annual Meeting of the Association for Computational Linguistics and the 11th International Joint Conference on Natural Language Processing (Volume 1: Long Papers)}, pages 340--350, Online. Association for Computational Linguistics.

\bibitem[{Cheung et~al.(2012)Cheung, Sia, and Kuan}]{cheung2012review}
Cindy Man-Yee Cheung, Choon-Ling Sia, and Kevin~KY Kuan. 2012.
\newblock Is this review believable? a study of factors affecting the credibility of online consumer reviews from an elm perspective.
\newblock \emph{Journal of the Association for Information Systems}, 13(8):2.

\bibitem[{Chu and Liu(2019)}]{chu2019meansum}
Eric Chu and Peter Liu. 2019.
\newblock Meansum: A neural model for unsupervised multi-document abstractive summarization.
\newblock In \emph{International Conference on Machine Learning}, pages 1223--1232. PMLR.

\bibitem[{Gao et~al.(2021)Gao, Yao, and Chen}]{gao2021simcse}
Tianyu Gao, Xingcheng Yao, and Danqi Chen. 2021.
\newblock {SimCSE}: Simple contrastive learning of sentence embeddings.
\newblock In \emph{Empirical Methods in Natural Language Processing (EMNLP)}.

\bibitem[{He et~al.(2020)He, Liu, Gao, and Chen}]{hedeberta}
Pengcheng He, Xiaodong Liu, Jianfeng Gao, and Weizhu Chen. 2020.
\newblock Deberta: Decoding-enhanced bert with disentangled attention.
\newblock In \emph{International Conference on Learning Representations}.

\bibitem[{Hosking et~al.(2023)Hosking, Tang, and Lapata}]{hosking-etal-2023-attributable}
Tom Hosking, Hao Tang, and Mirella Lapata. 2023.
\newblock Attributable and scalable opinion summarization.
\newblock In \emph{Proceedings of the 61st Annual Meeting of the Association for Computational Linguistics (Volume 1: Long Papers)}, pages 8488--8505, Toronto, Canada. Association for Computational Linguistics.

\bibitem[{Iso et~al.(2021)Iso, Wang, Suhara, Angelidis, and Tan}]{iso2021convex}
Hayate Iso, Xiaolan Wang, Yoshihiko Suhara, Stefanos Angelidis, and Wang-Chiew Tan. 2021.
\newblock Convex aggregation for opinion summarization.
\newblock In \emph{Findings of the Association for Computational Linguistics: EMNLP 2021}, pages 3885--3903.

\bibitem[{Isonuma et~al.(2021)Isonuma, Mori, Bollegala, and Sakata}]{isonuma2021unsupervised}
Masaru Isonuma, Junichiro Mori, Danushka Bollegala, and Ichiro Sakata. 2021.
\newblock Unsupervised abstractive opinion summarization by generating sentences with tree-structured topic guidance.
\newblock \emph{Transactions of the Association for Computational Linguistics}, 9:945--961.

\bibitem[{Kitaev and Klein(2018)}]{kitaev-klein-2018-constituency}
Nikita Kitaev and Dan Klein. 2018.
\newblock Constituency parsing with a self-attentive encoder.
\newblock In \emph{Proceedings of the 56th Annual Meeting of the Association for Computational Linguistics (Volume 1: Long Papers)}, pages 2676--2686, Melbourne, Australia. Association for Computational Linguistics.

\bibitem[{Ko et~al.(2019)Ko, Durrett, and Li}]{ko2019domain}
Wei-Jen Ko, Greg Durrett, and Junyi~Jessy Li. 2019.
\newblock Domain agnostic real-valued specificity prediction.
\newblock In \emph{Proceedings of the AAAI Conference on Artificial Intelligence}, volume~33, pages 6610--6617.

\bibitem[{Koehn(2004)}]{koehn-2004-statistical}
Philipp Koehn. 2004.
\newblock Statistical significance tests for machine translation evaluation.
\newblock In \emph{Proceedings of the 2004 Conference on Empirical Methods in Natural Language Processing}, pages 388--395, Barcelona, Spain. Association for Computational Linguistics.

\bibitem[{Laban et~al.(2022)Laban, Schnabel, Bennett, and Hearst}]{laban2022summac}
Philippe Laban, Tobias Schnabel, Paul~N Bennett, and Marti~A Hearst. 2022.
\newblock Summac: Re-visiting nli-based models for inconsistency detection in summarization.
\newblock \emph{Transactions of the Association for Computational Linguistics}, 10:163--177.

\bibitem[{Li et~al.(2023)Li, Chowdhury, and Chaturvedi}]{li2023aspect}
Haoyuan Li, Somnath Basu~Roy Chowdhury, and Snigdha Chaturvedi. 2023.
\newblock Aspect-aware unsupervised extractive opinion summarization.
\newblock In \emph{Findings of the Association for Computational Linguistics: ACL 2023}, pages 12662--12678.

\bibitem[{Liu et~al.(2019)Liu, Ott, Goyal, Du, Joshi, Chen, Levy, Lewis, Zettlemoyer, and Stoyanov}]{liu2019roberta}
Yinhan Liu, Myle Ott, Naman Goyal, Jingfei Du, Mandar Joshi, Danqi Chen, Omer Levy, Mike Lewis, Luke Zettlemoyer, and Veselin Stoyanov. 2019.
\newblock Roberta: A robustly optimized bert pretraining approach.
\newblock \emph{arXiv preprint arXiv:1907.11692}.

\bibitem[{Louis and Maynez(2023)}]{louis-maynez-2023-opinesum}
Annie Louis and Joshua Maynez. 2023.
\newblock {O}pine{S}um: Entailment-based self-training for abstractive opinion summarization.
\newblock In \emph{Findings of the Association for Computational Linguistics: ACL 2023}, pages 10774--10790, Toronto, Canada. Association for Computational Linguistics.

\bibitem[{Miao et~al.(2020)Miao, Li, Wang, and Tan}]{miao2020snippext}
Zhengjie Miao, Yuliang Li, Xiaolan Wang, and Wang-Chiew Tan. 2020.
\newblock Snippext: Semi-supervised opinion mining with augmented data.
\newblock In \emph{Proceedings of The Web Conference 2020}, pages 617--628.

\bibitem[{Ouyang et~al.(2022)Ouyang, Wu, Jiang, Almeida, Wainwright, Mishkin, Zhang, Agarwal, Slama, Ray et~al.}]{ouyang2022training}
Long Ouyang, Jeffrey Wu, Xu~Jiang, Diogo Almeida, Carroll Wainwright, Pamela Mishkin, Chong Zhang, Sandhini Agarwal, Katarina Slama, Alex Ray, et~al. 2022.
\newblock Training language models to follow instructions with human feedback.
\newblock \emph{Advances in Neural Information Processing Systems}, 35:27730--27744.

\bibitem[{Radford et~al.(2019)Radford, Wu, Child, Luan, Amodei, Sutskever et~al.}]{radford2019language}
Alec Radford, Jeffrey Wu, Rewon Child, David Luan, Dario Amodei, Ilya Sutskever, et~al. 2019.
\newblock Language models are unsupervised multitask learners.
\newblock \emph{OpenAI blog}, 1(8):9.

\bibitem[{{\v R}eh{\r u}{\v r}ek and Sojka(2010)}]{rehurek_lrec}
Radim {\v R}eh{\r u}{\v r}ek and Petr Sojka. 2010.
\newblock {Software Framework for Topic Modelling with Large Corpora}.
\newblock In \emph{{Proceedings of the LREC 2010 Workshop on New Challenges for NLP Frameworks}}, pages 45--50, Valletta, Malta. ELRA.

\bibitem[{Rousseeuw(1987)}]{rousseeuw1987silhouettes}
Peter~J Rousseeuw. 1987.
\newblock Silhouettes: a graphical aid to the interpretation and validation of cluster analysis.
\newblock \emph{Journal of computational and applied mathematics}, 20:53--65.

\bibitem[{Suhara et~al.(2020)Suhara, Wang, Angelidis, and Tan}]{suhara2020opiniondigest}
Yoshihiko Suhara, Xiaolan Wang, Stefanos Angelidis, and Wang-Chiew Tan. 2020.
\newblock Opiniondigest: A simple framework for opinion summarization.
\newblock In \emph{Proceedings of the 58th Annual Meeting of the Association for Computational Linguistics}, pages 5789--5798.

\bibitem[{Vaswani et~al.(2017)Vaswani, Shazeer, Parmar, Uszkoreit, Jones, Gomez, Kaiser, and Polosukhin}]{vaswani2017attention}
Ashish Vaswani, Noam Shazeer, Niki Parmar, Jakob Uszkoreit, Llion Jones, Aidan~N Gomez, {\L}ukasz Kaiser, and Illia Polosukhin. 2017.
\newblock Attention is all you need.
\newblock \emph{Advances in neural information processing systems}, 30.

\bibitem[{Williams et~al.(2018)Williams, Nangia, and Bowman}]{N18-1101}
Adina Williams, Nikita Nangia, and Samuel Bowman. 2018.
\newblock A broad-coverage challenge corpus for sentence understanding through inference.
\newblock In \emph{Proceedings of the 2018 Conference of the North American Chapter of the Association for Computational Linguistics: Human Language Technologies, Volume 1 (Long Papers)}, pages 1112--1122. Association for Computational Linguistics.

\bibitem[{Zhao and Chaturvedi(2020)}]{zhao2020weakly}
Chao Zhao and Snigdha Chaturvedi. 2020.
\newblock Weakly-supervised opinion summarization by leveraging external information.
\newblock In \emph{Proceedings of the AAAI Conference on Artificial Intelligence}, volume~34, pages 9644--9651.

\end{thebibliography}
\clearpage
\appendix
\section{Appendix}
\subsection{Text Segmentation}
\label{segment}

Due to the nature of reviews, many review sentences discuss several unrelated aspects, such as `The room is spacious and staff are helpful.' . These sentences might make rationales less specific to the representative opinions 
because they might contain unrelated information concerning a certain opinion. To alleviate these problems, \Ra extracts clauses from review sentences using a constituency parser \cite{kitaev-klein-2018-constituency}. The goal of the extraction is to reach a balance of two criteria. First, the resulting clauses are complete and fluent sentences. Second, the most resulting clauses only discuss one aspect. 

\begin{figure*}
     \centering
     \begin{subfigure}[t]{0.32\textwidth}
         \centering
         \includegraphics[width=\textwidth]{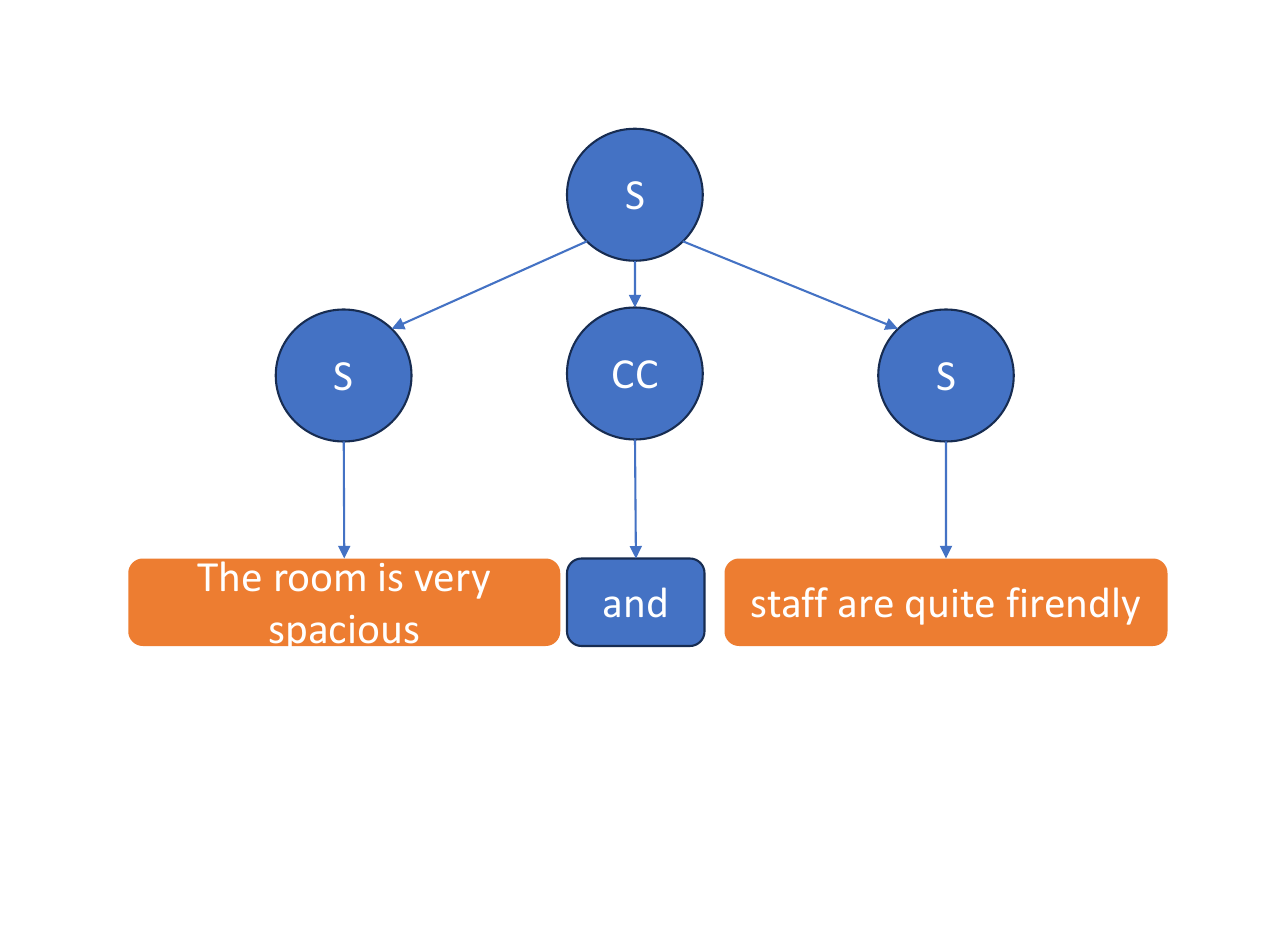}
         \caption{The root has two clause children and therefore two corresponding clauses are extracted from the sentence. }
         \label{fig:tree1}
     \end{subfigure}
     \hfill
     \begin{subfigure}[t]{0.32\textwidth}
         \centering
         \includegraphics[width=\textwidth]{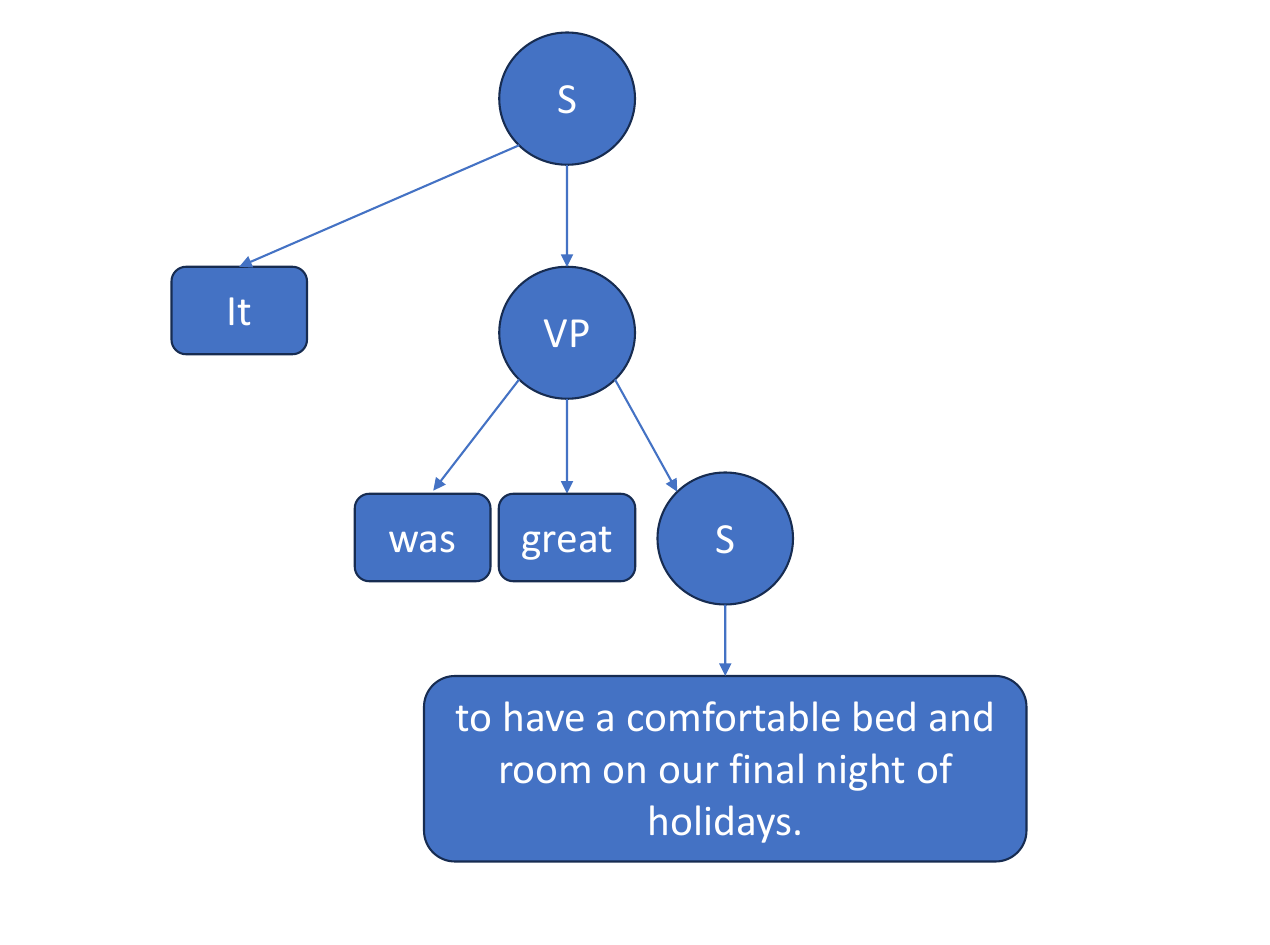}
         \caption{The whole sentence is extracted to keep the information complete since there is only one clause in the sentence.}
         \label{fig:tree2}
     \end{subfigure}
     \hfill
     \begin{subfigure}[t]{0.32\textwidth}
         \centering
         \includegraphics[width=\textwidth]{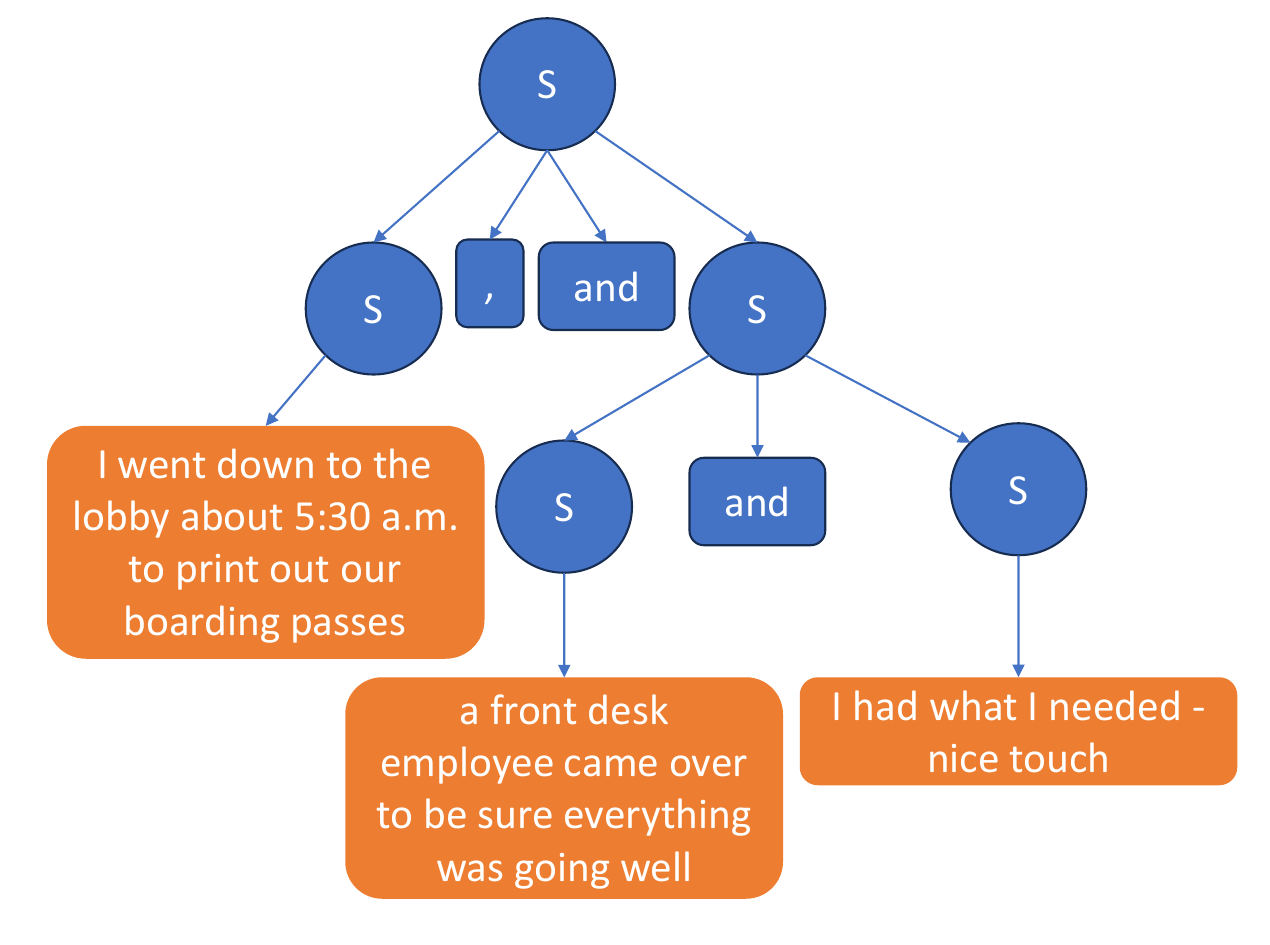}
         \caption{The root has two children clause. Since the length of the second clause is longer than the maximum length, \Ra traverse its children and find two children clause. Therefore, three clause are extracted from the sentence.}
         \label{fig:tree3}
     \end{subfigure}
        \caption{Three sentences and their constituency parsing trees. A orange box denotes one extracted clause. }
        \label{fig:threetree}
\end{figure*}

Given a parse tree of a sentence, \Ra traverse it from its root to determine the boundary of a clause. When a node whose tag is `S' is traversed, if it has not been extracted yet, \Ra will check the length of the corresponding clause. For clauses longer than the maximum length $\epsilon$, they still might discuss several aspects. Therefore, \Ra further traverse all their children as in Fig. \ref{fig:tree3}. For clauses shorter than the minimum length $\gamma$, the clauses might be incomplete and the traversal stops at these nodes. The traversal also stops at the node whose tag is `SBAR' since the corresponding clauses usually complement other clauses. If the length of the corresponding clause is between the maximum length $\epsilon$ and $\gamma$, \Ra will extract the clause as in Fig. \ref{fig:tree1}. If \Ra only extracts one clause from the sentence, \Ra will extract the whole sentence instead to keep the information complete as in Fig. \ref{fig:tree2}. If \Ra extracts more than one clause from the sentence, \Ra will further check the distances between neighboring clauses. If the distance between any two neighboring clauses is larger than $\gamma$, \Ra will also extract the whole sentence. Otherwise, \Ra extracts the clauses. The above process extracts as many clauses as possible while keeping the extracting clauses complete. In the experiment, we set the maximum length as 20 and minimum length as 2.

\subsection{Preprocessing of Yelp Dataset}
\label{sec:prep_yelp}
For yelp dataset, we remove entities that contain less than 20 reviews. For entities containing more than 200 reviews, we randomly sample 200 reviews and discard other reviews of the entities to prevent dominant influences of some entities. For the remaining entities, we perform downsampling to create a dataset containing around 14.9K entities and around 1.22M reviews. The statistics are show in Table \ref{tab:dataset_stat}.

\begin{table}[t!]
\centering
\small
\begin{tabular}{l c c c} 
\toprule
Dataset & Train  & Dev. & Test  \TBstrut\\ 
\midrule
Space  & 11.2K/1.10M      & 50/5K        & 250/25K \Tstrut\\
Yelp & 14.9K/1.22M       & 50/5K       & 250/25K \Bstrut\\
\bottomrule
\end{tabular}
\caption{Dataset statistics for Space and Yelp. We report entity/review for each split of two datasets.}
\vspace{-3pt}
\label{tab:dataset_stat}
\end{table}

\subsection{Preprocessing for Keyword Extraction}
In \S \ref{sec:ration} and \S \ref{sec:ent_eval}, we extract keywords to evaluate the performance of \Ra. For this purpose, we perform the standard preprocessing. We first remove stop words using NLTK \cite{bird2009natural} and filter out extreme words using Gensim \cite{rehurek_lrec}. We finally perform lemmatization using NLTK. We show examples of extracted keywords in Table \ref{tab:keyword}.

\begin{table}[t]
\resizebox{0.45\textwidth}{!}{
\begin{tabular}{cc}
\hline
Opinion Cluster                                                                 & Keyword                                                                         \\ \hline
location is great                                                               & \begin{tabular}[c]{@{}c@{}}seattle downtown vintage\\ library walk\end{tabular} \\ \hline
\begin{tabular}[c]{@{}c@{}}bed is super comfortable\\ bed is great\end{tabular} & \begin{tabular}[c]{@{}c@{}}pillow linen comfy\\ ever mattress\end{tabular}     \\ \hline
\end{tabular}}
\caption{Samples of opinion clusters and keywords extracted from their rationale candidates. Keywords are shown in descending order of TF-IDF. Most keywords represent details highly related to but not repetitive of the corresponding opinion groups.}
\label{tab:keyword}
\end{table}

\subsection{Instruction for InstructGPT}
\label{sec:instruct}
\begin{figure*}
\centering
\includegraphics[width=0.95\textwidth]{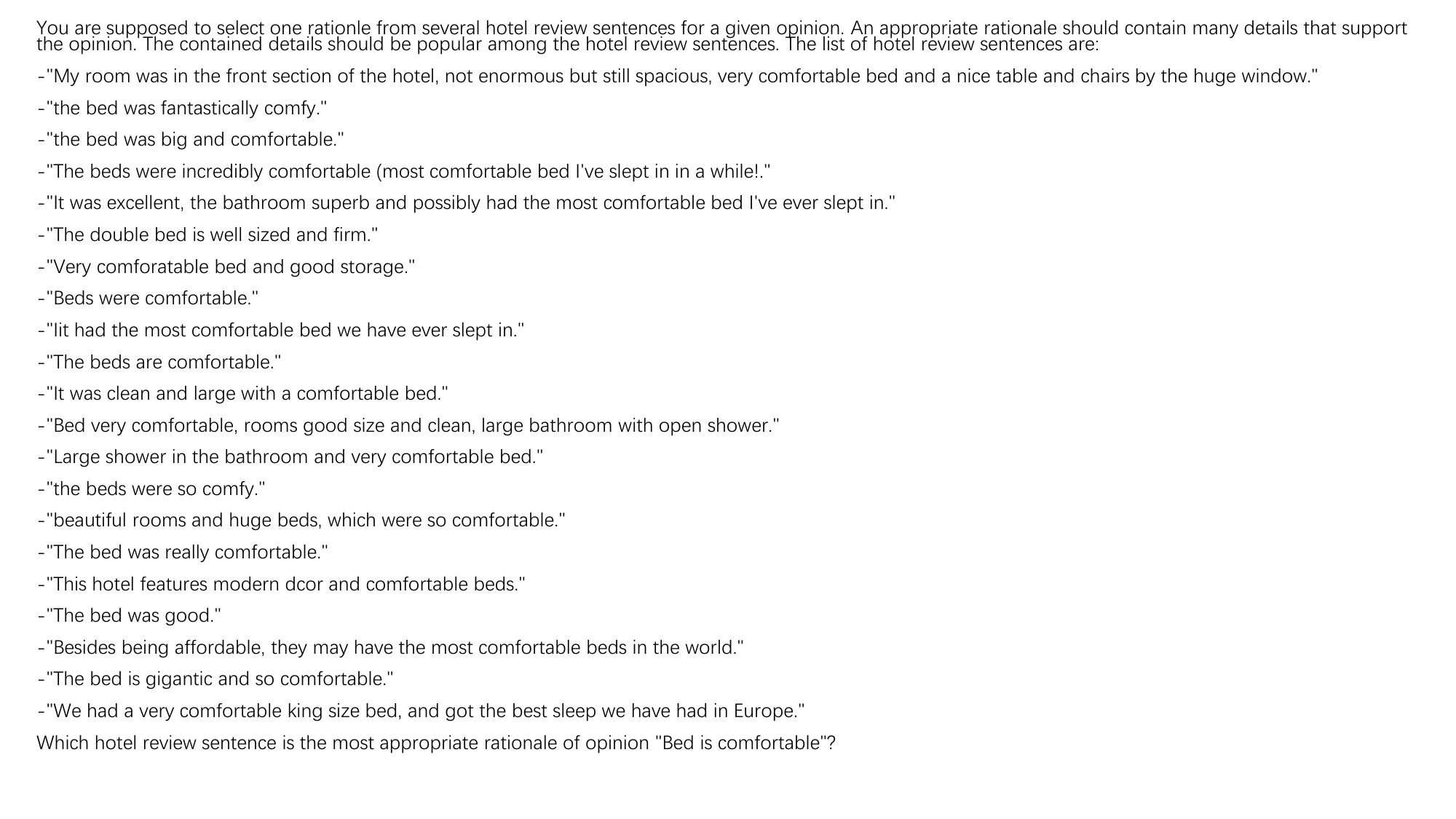}
\caption{Example instruction for extracting one rationale for each representative opinion using InstructGPT.}
\label{fig:oneration}
\end{figure*}

\begin{figure*}
\centering
\includegraphics[width=0.95\textwidth]{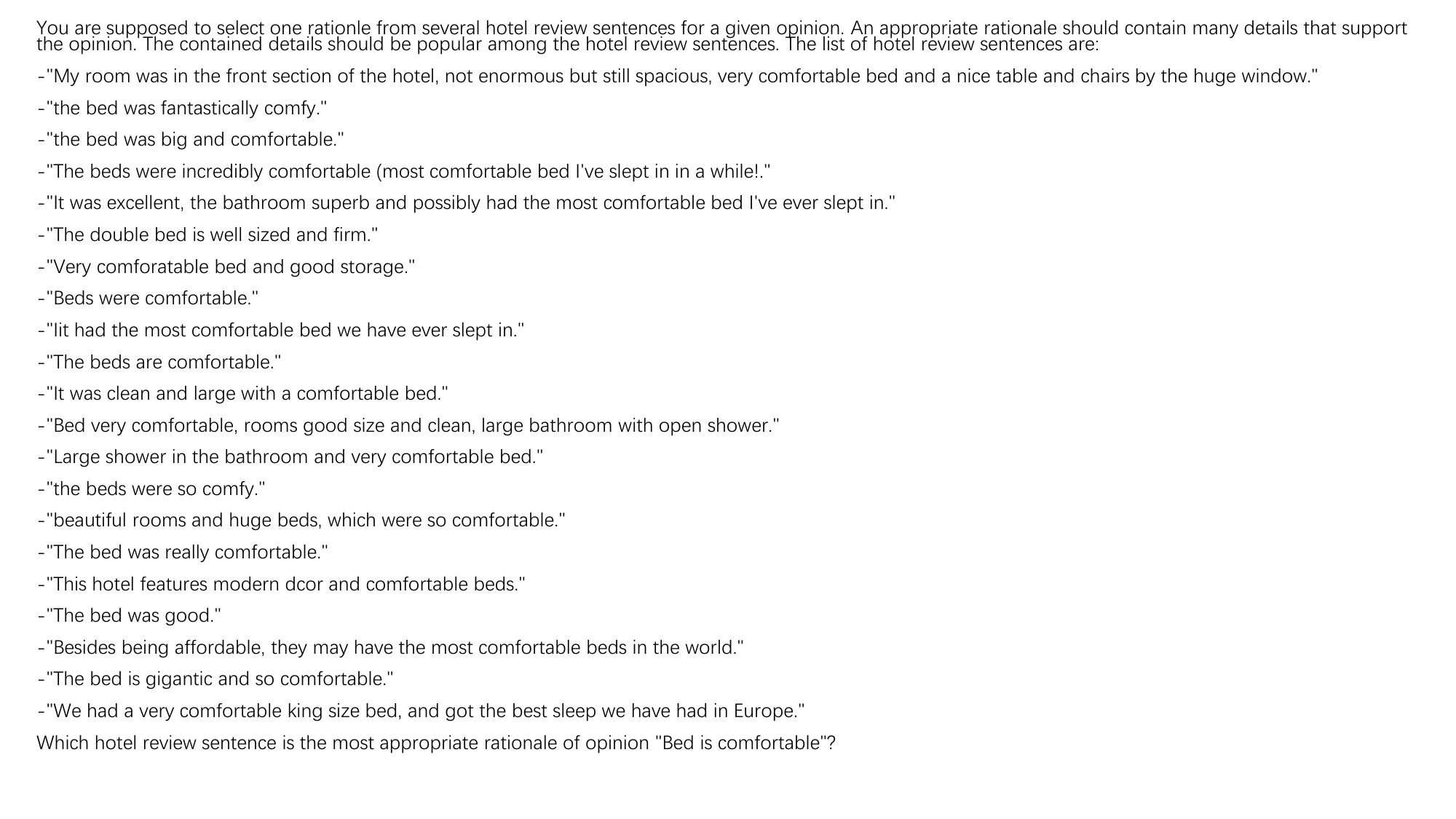}
\caption{Example instruction for extracting three rationales for each representative opinion using InstructGPT.}
\label{fig:threeration}
\end{figure*}

To extract rationales using InstructGPT, we provide the instructions that describe the four desirable properties of rationales as well as the representative opinion and its corresponding rationales. Under the extractive setting, we try several variations of prompts including paraphrasing, reordering, and restructuring the instruction material. We show the best instruction that we use for extracting one rationales for each opinion in Figure \ref{fig:oneration} and extracting three rationale for each opinion in Figure \ref{fig:threeration}.  
\begin{figure*}
\centering
\includegraphics[width=0.95\textwidth]{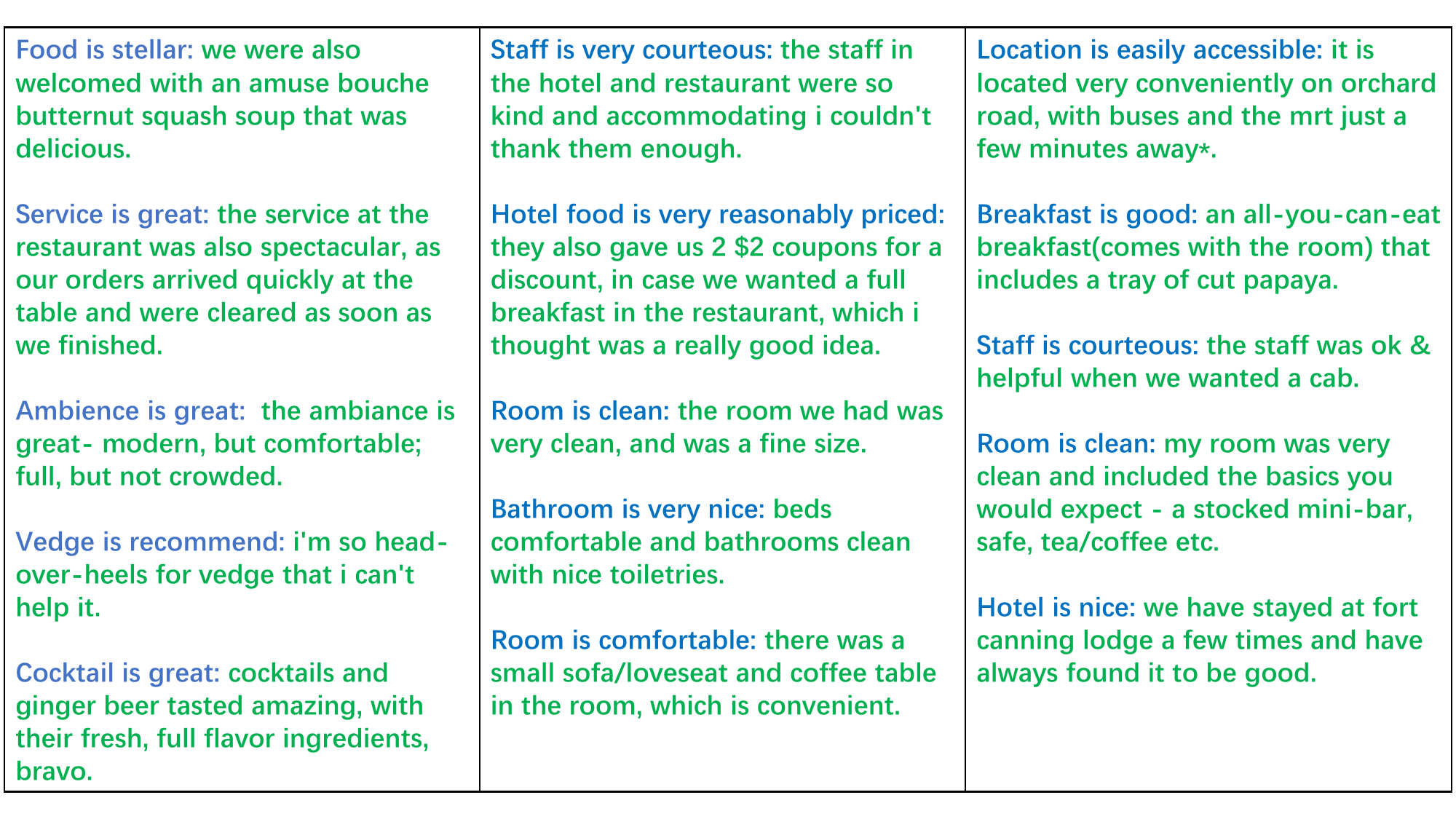}
\caption{Three sample rationale-based summaries. Each line presents a \textcolor[RGB]{0,112,192}{representative opinion} and its \textcolor[RGB]{0,176,80}{rationale}.}
\label{fig:sample}
\end{figure*}
\begin{figure*}
\centering
\includegraphics[width=0.95\textwidth]{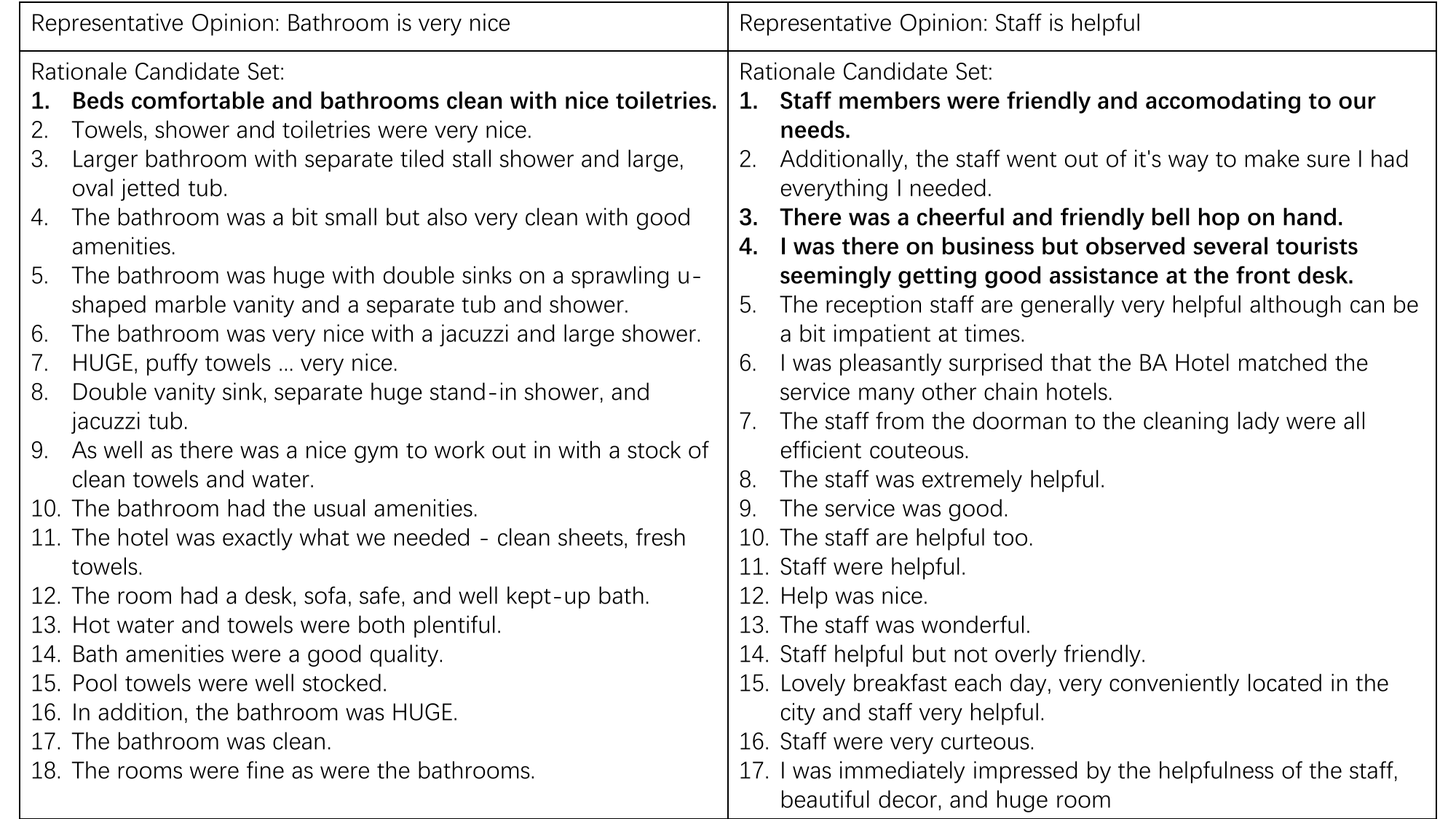}
\caption{Rationale candidate sets of sample representative opinions on the Space dataset. \textbf{Bold} sentences are extracted as rationales when one rationale (left) or three rationales (right) are extracted. }
\label{fig:sample_rationale_candidate1}
\end{figure*}
\begin{figure*}
\centering
\includegraphics[width=0.95\textwidth]{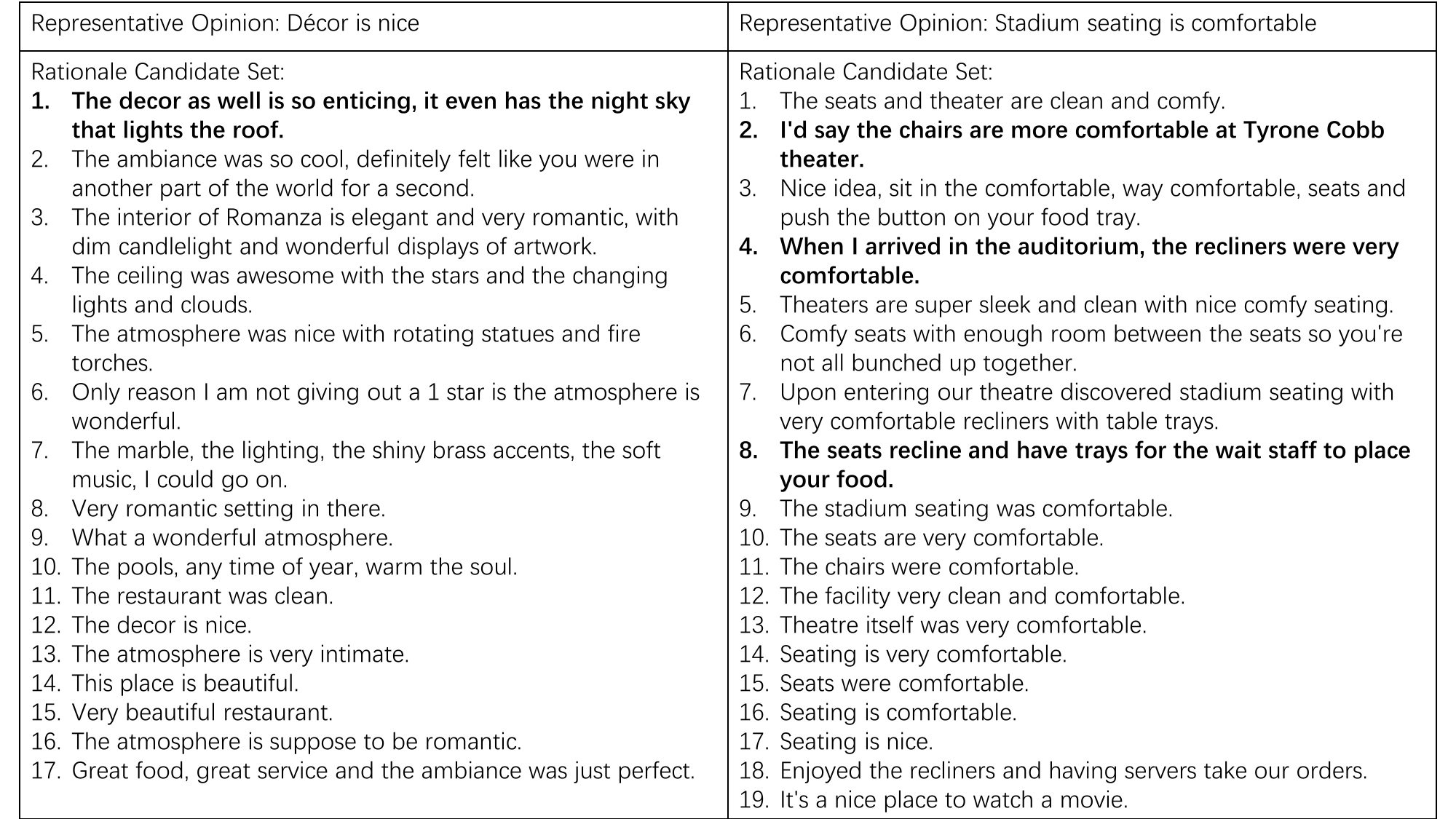}
\caption{Rationale candidate sets of sample representative opinions on the Yelp dataset. \textbf{Bold} sentences are extracted as rationales when one rationale (left) or three rationales (right) are extracted. }
\label{fig:sample_rationale_candidate2}
\end{figure*}
\begin{figure*}
\centering
\includegraphics[width=0.8\textwidth]{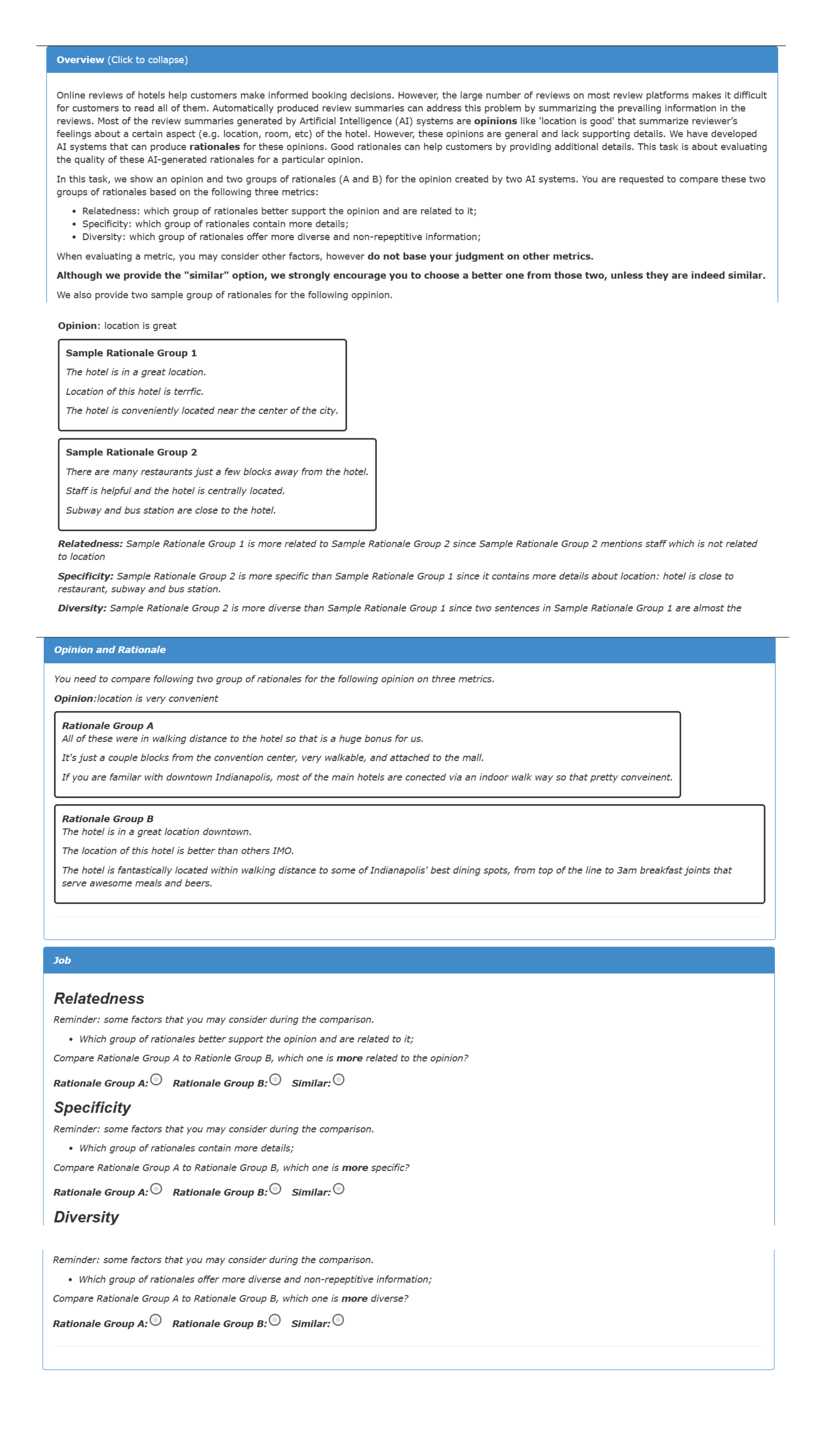}
\caption{AMT instructions for human evaluation for comparing rationales.}
\label{fig:human_eval}
\end{figure*}
\begin{figure*}
\centering
\includegraphics[width=0.8\textwidth]{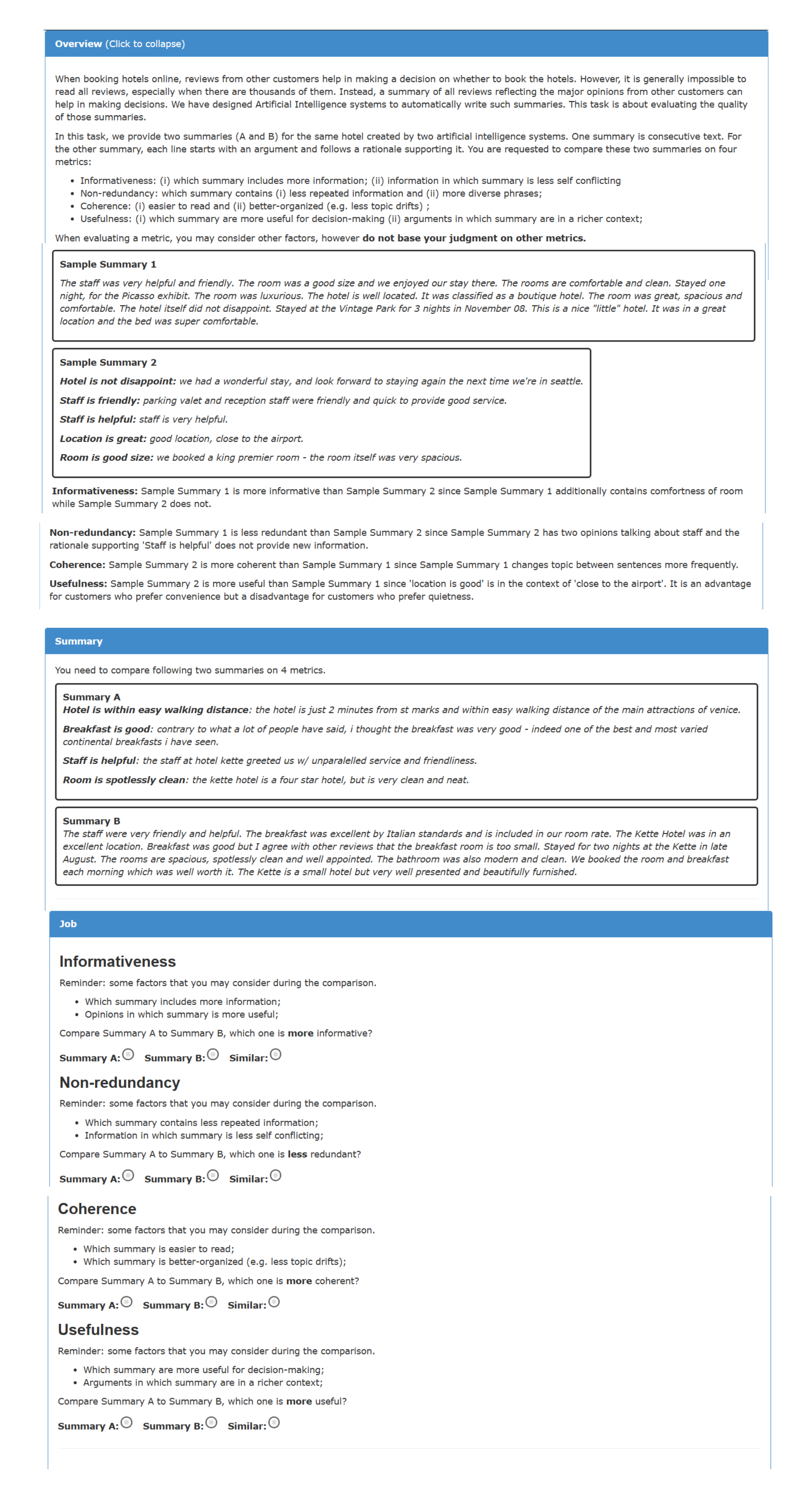}
\caption{AMT instructions for human evaluation of comparing summaries.}
\label{fig:human_eval2}
\end{figure*}

\subsection{Error Analysis of InstructGPT Rationale}
\label{sec:instruct_rationale}
 As discussed in Section \ref{auto}, the performance of InstructGPT was encouraging for an initial study but not up to the mark. Specifically, we manually analyzed the InstructGPT’s rationales while also asking for explanations of those rationales. We found that the rationales extracted by the InstructGPT were lacking in many senses. First,  the InstructGPT might ignore the part of the instruction that required the rationales to have additional details as compared to the opinions. The extracted rationales were simply paraphrases of the opinions. This defeats the purpose of having rationales. Second, the InstructGPT might misunderstand what is meant by “containing additional details”. For example, it might focus too much on plural forms or tenses of certain words. The InstructGPT might think “Rooms are great” is an appropriate rationale for “Room is great” because maybe “Rooms are great” suggests there are many rooms that are great instead of one room. We faced these problems even after trying multiple prompts. In the future, as LLMs hopefully improve, future works could revisit this problem for better solutions.

\subsection{Human Evaluation}
The human annotators are required to be in the United States, have HIT Approval Rate greater than 98, and be masters. The screenshot of the human evaluation interface for rationale evaluation is shown in Figure \ref{fig:human_eval}. The screenshot of the human evaluation interface for summary evaluation is shown in Figure \ref{fig:human_eval2}.

\subsection{Implementation Detail of Rationale Candidate Set Evaluation }
\label{sec:detail_candidate}
We compare \Ra with four baseline models: $\rm RoBERta_{mnli}$, $\rm KPA$, $\rm Snippext$, and $\rm Hercules$. $\rm RoBERta_{mnli}$ uses RoBERta-large \cite{liu2019roberta} finetuned on the MNLI dataset \cite{N18-1101}. $\rm RoBERta_{mnli}$ then uses the finetuned model to estimate the alignment probability between review sentences and representative opinions

$\rm KPA$ uses RoBERta-large \cite{liu2019roberta} as the base model and performs the same domain adaptation as \Ra. The model is then finetuned on ArgKP dataset using the same hyperparameters as \cite{bar-haim-etal-2021-every}. $\rm KPA$ then uses the finetuned model to estimate the alignment probability. 

$\rm Snippext$ uses the ABSA model \cite{miao2020snippext}. $\rm Snippext$ estimates the aspect distribution and the sentiment distribution for each representative opinion and review sentence based on the ABSA model. $\rm Snippext$ then estimates the alignment probability as the product of the cosine similarity of aspect distribution and the sentiment distribution between representaitive opinions and review sentences. 

$\rm Hercules$ generates the path representation on a tree for each representative opinions and review sentences. If a representative opinion and a review sentence has the same first node of their path representation, the review sentence belongs to the rationale candidate set of that representative opinion. 

For a fair comparison, we extract an average of $8$ rationale candidate sets for each entity and all rationale candidate sets on average cover $30\%$ of review sentences except for $\rm Hercules$. When using SimCSE to obtain sentence representations, we use `unsup-simcse-roberta-large' version.

\subsection{Experiment with Hercules}
\label{sec:hercules_experiment}
\Ra is independent of the choice of extractive summarization systems and can work with other extractive summarization systems. In this section, we show the automatic metrics on the Space dataset when the representative opinions are extracted from the summaries produced by the extractive version of $\rm Hercules$. All the implementation details are the same. 

We show the automatic metric for evaulating rationale candidate sets in Table \ref{tab:setevalher}. 
\begin{table}[t]
\centering
\resizebox{0.48\textwidth}{!}{
\begin{tabular}{lcccc}
\hline
         & \textit{Silh}     & \textit{NPMI}           & \textit{SC}           & Overall  \\ \hline
         & \multicolumn{4}{c}{Space}                           \\
$\rm RoBERta_{mnli}$   & 0.088 & -0.034 & 0.953 & 0.693 \\
$\rm KPA$      & 0.142 & -0.013 & 0.946 & 0.925 \\
$\rm Snippext$     & 0.122 & -0.065 & 0.916 & 0.270 \\
$\rm Hercules$ & 0.009 & -0.064 & 0.944 & 0.263 \\
\Ra   & 0.135 & -0.012 & 0.952 & \textbf{0.974} \\ \hline
\end{tabular}}
\caption{Automatic evaluation of rationale candidate sets when the representative opinions are extracted from summaries generated by $\rm Hercules$. Considering the three measures and their overall scores, \Ra still generates rationale candidates of better quality than the baselines when using the other extractive opinion summarization system. }
\label{tab:setevalher}
\end{table}
It can be observed that \Ra still generates rationale candidates of better quality than the baselines when using the other extractive opinion summarization system, which also shows \Ra is independent of extractive summarization systems. 

We show the automatic metrics for evaluating rationales in Table \ref{tab:autoher}. 
\begin{table}[t]
\centering 
\resizebox{0.47\textwidth}{!}{
\begin{tabular}{lccccc}
\hline
            & $emb_{rel}$ & $key_{spec}$ & $key_{pop}$ & $emb_{div}$ & Overall \\ \hline
            & \multicolumn{5}{c}{Space (k=1)}             \\
\Ra          &   0.399      & 0.236     & 0.237    &    -    & 0.728  \\
~~~w/o $rel$      & 0.400        & 0.239     & 0.237    &  -    & \textbf{0.748}    \\
~~~w/o $spec$     & 0.525        & 0.174     & 0.219    &   -   & 0.554   \\
~~~w/o $pop$       & 0.349         & 0.212     & 0.207    &  -   &  0.389   \\
InstructGPT &  0.558        & 0.167     & 0.172    &   -    &  0.333 \\  \hdashline
            & \multicolumn{5}{c}{Space (k=3)}             \\
\Ra          &   0.390       & 0.520     & 0.239    &  0.577   &    \textbf{0.623} \\
~~~w/o $rel$       & 0.391         & 0.524     & 0.238    &  0.572   &  0.614   \\
~~~w/o $spec$       &  0.474        & 0.456     & 0.240    & 0.546   & 0.485     \\
~~~w/o $pop$        &  0.351        & 0.496     & 0.222    &  0.631  &  0.399    \\
~~~w/o $div$       &   0.398       & 0.511     & 0.241    &   0.555   & 0.575    \\  \hline
\end{tabular}}
\caption{Automatic evaluation of rationales on the Space dataset with one (k=1) and three (k=3) rationales extracted per representative opinion. Considering the four measures and their \textit{overall} values, \Ra still extracts the best rationales when the representative opinions are extracted from summaries generated by the other extractive summarization system. }
\label{tab:autoher}
\end{table}
It can be observed that RATION also extracts the best rationales when the representative opinions are extracted from the summaries produced by Hercules.
\subsection{Error Analysis}
\label{sec:error}
\Ra occasionally generates undesirable rationale-based opinion summaries. We analyze these summaries and find the most common errors are the extracted rationales of an opinion not containing many related details of that opinion. For example, in the right sample of Figure \ref{fig:sample}, the extracted rationale for the opinion `Room is clean', `my room was very clean and included the basics you would expect - a stocked mini-bar, safe, tea/coffe etc.', only mentions `clean' and contains lots of details not related to the detail. The main reason is that \Ra separately estimates the specificity and relatedness as mentioned in Section \ref{sec:limit}. Suppose a sentence discusses aspect X and aspect Y, and it only briefly mentions X but contains lots of details related to Y. When extracting rationales for an opinion about aspects X, the sentence would have a high relatedness score because it mentions X. It would also have a high specificity score because it contains many details. We reduce such errors by dividing review sentences into clauses and extracting clauses as rationales (Appendix \ref{segment}). However, some resulting clauses might still discuss multiple aspects. Future work can explore how to jointly model these four properties at the same time. 

We also find other less frequent errors, such as some representative opinions being too similar and the alignment model making wrong estimations.
\end{document}